\batchmode
\makeatletter
\def\input@path{{N:/Gemeinsam/Marktrisiko/Flaig/DataScientist/Promotion/Promotionsdateien/}}
\makeatother
\documentclass[ngerman,english,british]{article}
\usepackage[T1]{fontenc}
\usepackage[latin9]{inputenc}
\usepackage{geometry}
\usepackage{babel}
\usepackage{array}
\usepackage{verbatim}
\usepackage{longtable}
\usepackage{float}
\usepackage{booktabs}
\usepackage{textcomp}
\usepackage{amsmath}
\usepackage{amsthm}
\usepackage{amssymb}
\usepackage{graphicx}
\usepackage{wasysym}
\usepackage[authoryear]{natbib}
\usepackage[pdftex,unicode=true,pdfusetitle,
 bookmarks=true,bookmarksnumbered=false,bookmarksopen=false,
 breaklinks=false,pdfborder={0 0 0},pdfborderstyle={},backref=false,colorlinks=false]
 {hyperref}

\makeatletter

\providecommand{\tabularnewline}{\\}
\floatstyle{ruled}
\newfloat{algorithm}{tbp}{loa}
\providecommand{\algorithmname}{Algorithm}
\floatname{algorithm}{\protect\algorithmname}

\numberwithin{equation}{section}
\numberwithin{figure}{section}
\newenvironment{lyxcode}
	{\par\begin{list}{}{
		\setlength{\rightmargin}{\leftmargin}
		\setlength{\listparindent}{0pt}
		\raggedright
		\setlength{\itemsep}{0pt}
		\setlength{\parsep}{0pt}
		\normalfont\ttfamily}%
	 \item[]}
	{\end{list}}
\theoremstyle{plain}
\ifx\thechapter\undefined
	\newtheorem{thm}{\protect\theoremname}
\else
	\newtheorem{thm}{\protect\theoremname}[chapter]
\fi
\theoremstyle{definition}
\newtheorem{defn}[thm]{\protect\definitionname}

\usepackage[titles]{tocloft}

\usepackage{url}

\makeatother

\addto\captionsbritish{\renewcommand{\definitionname}{Definition}}
\addto\captionsbritish{\renewcommand{\theoremname}{Theorem}}
\addto\captionsenglish{\renewcommand{\algorithmname}{Algorithm}}
\addto\captionsenglish{\renewcommand{\definitionname}{Definition}}
\addto\captionsenglish{\renewcommand{\theoremname}{Theorem}}
\addto\captionsngerman{\renewcommand{\algorithmname}{Algorithmus}}
\addto\captionsngerman{\renewcommand{\definitionname}{Definition}}
\addto\captionsngerman{\renewcommand{\theoremname}{Theorem}}
\providecommand{\definitionname}{Definition}
\providecommand{\theoremname}{Theorem}

\begin{document}
\title{Scenario generation for market risk models using generative neural
networks}
\author{Solveig Flaig\thanks{Corresponding author. Deutsche Rueckversicherung AG, Kapitalanlage
/ Market risk management, Hansaallee 177, 40549 Duesseldorf, Germany.
E-Mail: solveig.flaig@deutscherueck.de.} $\:$\thanks{Carl von Ossietzky Universität, Institut für Mathematik, 26111 Oldenburg,
Germany.} , Gero Junike\thanks{Carl von Ossietzky Universität, Institut für Mathematik, 26111 Oldenburg,
Germany. E-Mail: gero.junike@uol.de.}}
\date{07.09.2022}
\maketitle
\begin{abstract}
In this research, we show how existing approaches of using generative
adversarial networks (GANs) as economic scenario generators (ESG)
can be extended to a whole internal market risk model - with enough
risk factors to model the full band-width of investments for an insurance
company and for a time horizon of one year as required in Solvency
2. We demonstrate that the results of a GAN-based internal model are
similar to regulatory approved internal models in Europe. Therefore,
GAN-based models can be seen as a data-driven alternative way of market
risk modeling.
\end{abstract}
\textbf{\emph{}}
\begin{lyxcode}
\global\long\def\E{\mathbb{E}}%

\global\long\def\G{\mathbb{G}}%

\global\long\def\R{\mathbb{R}}%

\global\long\def\N{\mathbb{N}}%

\global\long\def\P{\mathbb{\mathbb{P}}}%
\end{lyxcode}

\section{Introduction}

Generating realistic scenarios of how the financial markets might
behave in the future is one key component of internal market risk
models used by insurance companies for Solvency 2 purposes. Currently,
these are built using economic scenario generators (ESGs) based mainly
on financial mathematical models, see \citet{bennemann2011handbuch}
and \citet{pfeifer2018generating}. These ESGs require strong assumptions
on the behavior of the risk factors and their dependencies, are time-consuming
to calibrate and it is difficult in this framework to model complex
dependencies.

An alternative method for scenario generation can be a special type
of neural networks called generative adversarial networks (GANs),
invented by \citet{goodfellow2014pouget}. This network architecture
consists of two neural networks which has gained a lot of attention
due to its ability to generate realistic looking images, see \citet{aggarwal2021generative}.

As financial data, at least for liquid instruments, is consistently
available, GANs are used in various areas of finance, including market
prediction, tuning of trading models, portfolio management and optimization,
synthetic data generation and diverse types of fraud detection, see
\citet{eckerli2021generative}. \citet{henry2019generative}, \citet{lezmi2020improving},
\citet{fu2019time}, \citet{wiese2019deep}, \citet{ni2020conditional}
and \citet{wiese2020quant} have already used GANs for scenario generation
in the financial sector. The focus of their research was the generation
of financial time series for a limited number of risk factors (up
to 6) or a single asset class. To the best of our knowledge, there
is no research performing a full value-at-risk calculation for an
insurance portfolio based on GAN generated scenarios.

In this work, we perform a market risk calculation for typical insurance
portfolios using a GAN instead of a classical ESG. We base our research
on publicly available financial data from Bloomberg. The research
of \citet{ngwenduna2021alleviating}, \citet{cote2020synthesizing}
and \citet{kuo2019generative} also uses a GAN in an actuarial context,
but they use it to generate publicly available data from restricted
data or to create new data for solving the issue of having imbalanced
data sets. However, the methods introduced in those papers could be
used when dealing with illiquid instruments where no consistent tabular
data is available.

In this research we
\begin{itemize}
\item expand the scenario generation by a GAN to a complete market risk
calculation serving for Solvency 2 purposes in insurance companies
and
\item compare the results of a GAN-based ESG to the ESG approaches implemented
in regulatory approved market risk models in Europe.
\end{itemize}
As a novelty, this research shows that there is an alternative way
of market risk modeling beyond traditional ESGs which can also serve
regulatory approved models as they perform well in the EIOPA (European
Insurance and Occupational Pensions Authority) benchmarking study.
Therefore, the proof of concept of whether a GAN can serve as an ESG
for market risk modeling is successful.

The paper is structured as follows: In Section 2, we provide some
background both on market risk calculation under Solvency 2 and on
GANs. The MCRCS (market and credit risk comparison study), a benchmarking
exercise for approved market risk models in Europe conducted annually
by EIOPA, is also introduced in Section 2. Section 3 explains how
GANs can be used as ESGs, how they can be included in an internal
model process and how GANs can be implemented. Comparison of the results
of a GAN-based internal model with the results of the models in the
MCRCS study is presented in Section 4. A discussion of the stability
of these results is also included in this section. Section 5 concludes
and provides on overview of the differences of this GAN-based approach
with traditional methods.

\section{Background}

Before we present our work, we give a short introduction to the two
main topics involved: economic scenario generators (ESG) and their
usage for market risk calculation under Solvency 2 and generative
adversarial networks (GANs).

\subsection{Market risk calculation under Solvency 2}

In 2016, a new regulation for insurance companies in Europe was introduced:
Solvency 2. One central requirement is the calculation of the solvency
capital requirement, called SCR. The amount of SCR depends on the
risks to which the insurance company is exposed, see e.g. \citet[Chapter 4]{grundl2019solvency}.
The eligible capital of an insurance company is then compared with
the SCR to determine whether the eligible capital is sufficient to
cover all the risks taken by the insurance company.

The solvency capital requirement equals the Value-at-Risk (VaR) at
a 99.5\%-level for a time horizon of one year, see \citet[Chapter 2.3]{bennemann2011handbuch}.
A mathematical definition of the VaR and a derivation of its usage
in this context can be found in \citet[p. 69]{denuit2006actuarial}.

The risk of an insurer can be divided into six different modules:
market risk, health underwriting risk, counterparty default risk,
life underwriting risk, non-life underwriting risk and operational
risk. The modules themselves consist of sub-modules, see \citet{eiopa2014underlying}.
Market risk, e.g. consists of the six submodules interest rate, equity,
property, spread, currency and concentration risk.

The SCR can be calculated using either the standard model or an internal
model. For the standard model, the regulatory framework sets specific
rules for the calculation for each risk encountered by the insurance
company, defined in \citet{delegated2015commission}. Each internal
model has to cover the same types of risks as the standard model and
must be approved by local supervisors to ensure accordance with the
principles of Solvency 2.

In this work, we will focus on the calculation of the market risk
of a non-life insurer. However, the methods presented here can be
applied for other risks, too. The reason for selecting market risk
here is threefold:
\begin{itemize}
\item the underlying data in the financial market is publicly available
and equal for all insurers,
\item market risk forms a major part of the SCR of an insurance company
(\citet[p. 22]{eiopa2021insuranceoverview} states that market risk
accounts for 53\% of the net solvency capital requirement before diversification
benefits; this varies between life (59\%) and non-life (43\%) insurers)
and
\item a comprehensive benchmark exercise, called ``market and credit risk
comparison study'' MCRCS conducted by EIOPA is available for comparison
of the results.
\end{itemize}
Current internal models for market risk often use Monte-Carlo simulation
techniques to derive the risk of the (sub)modules and then use correlations
or copulas for aggregation, see \citet[p. 189]{bennemann2011handbuch}
and \citet{pfeifer2018generating}. The basis of the Monte-Carlo simulation
is a scenario generation performed by an ``economic scenario generator''
ESG.

A definition of an ESG can be found in \citet[p. 7]{pedersen2016economic}:
\begin{defn}
An economic scenario generator (ESG) is a computer-based model of
an economic environment that is used to produce simulations of the
joint behavior of financial market values and economic variables.
\end{defn}

The ESG implements financial-mathematical models for all relevant
risk factors (e.g. interest rate, equity) and their dependencies.
Under those scenarios, the investment and liabilities portfolio of
the insurer is evaluated and the risk is given by the 0.5\%-percentile
of the loss in these scenarios.

\subsection{Introduction to the MCRCS study}

Since 2017, EIOPA performs an annual study, called the \emph{market
and credit risk comparison study}, abbr. MCRCS. According to the instructions
from \citet{EIOPA_MCRCS_Instructions}, the \textquotedbl primary
objective of the MCRCS is to compare market and credit risk model
outputs for a set of realistic asset portfolios\textquotedbl . In
the study, all insurance undertakings with significant exposure in
EUR and with an approved internal model are asked to participate,
see \citet{EIOPA_MCRCS_Ergebnis}. In the study as of year-end 2019,
21 insurance companies from 8 different countries of the European
Union participated.

All participants have to model the risk of 104 different synthetic
instruments. Those comprise all relevant asset classes, i.e. risk-free
interest rates, sovereign bonds, corporate bonds, equity indices,
property, foreign exchange and some derivatives. A detailed overview
of the synthetic instruments that are used in this study can be found
in \citet{EIOPA_MCRCS_Excel}.

Additionally, those instruments are grouped into ten different asset-only
benchmark portfolios, two liability-only benchmark portfolios and
ten combined portfolios. These portfolios \textquotedbl should reflect
typical asset risk profiles of European insurance undertakings\textquotedbl ,
see \citet[Section 2]{EIOPA_MCRCS_Instructions}. This analysis sheds
light into the interaction and dependencies between the risk factors.

These 22 portfolios are denoted as follows:
\begin{itemize}
\item asset-only benchmark portfolios: BMP1, BMP2, ..., BMP10
\item liability-only benchmark portfolios: L1 and L2
\item combined portfolios: BMP1+L1, BMP3+L1, BMP7+L1, BMP9+L1, BMP10+L1,
BMP1+L2, BMP3+L2, BMP7+L2, BMP9+L2, BMP10+L2.
\end{itemize}
The combined portfolios are linear combinations of the asset and liability
benchmark portfolios, e.g. BMP1+L1 combines the asset-only benchmark
portfolio BMP1 with liability portfolio L1.

\begin{figure}[H]
\begin{centering}
\includegraphics[width=10cm]{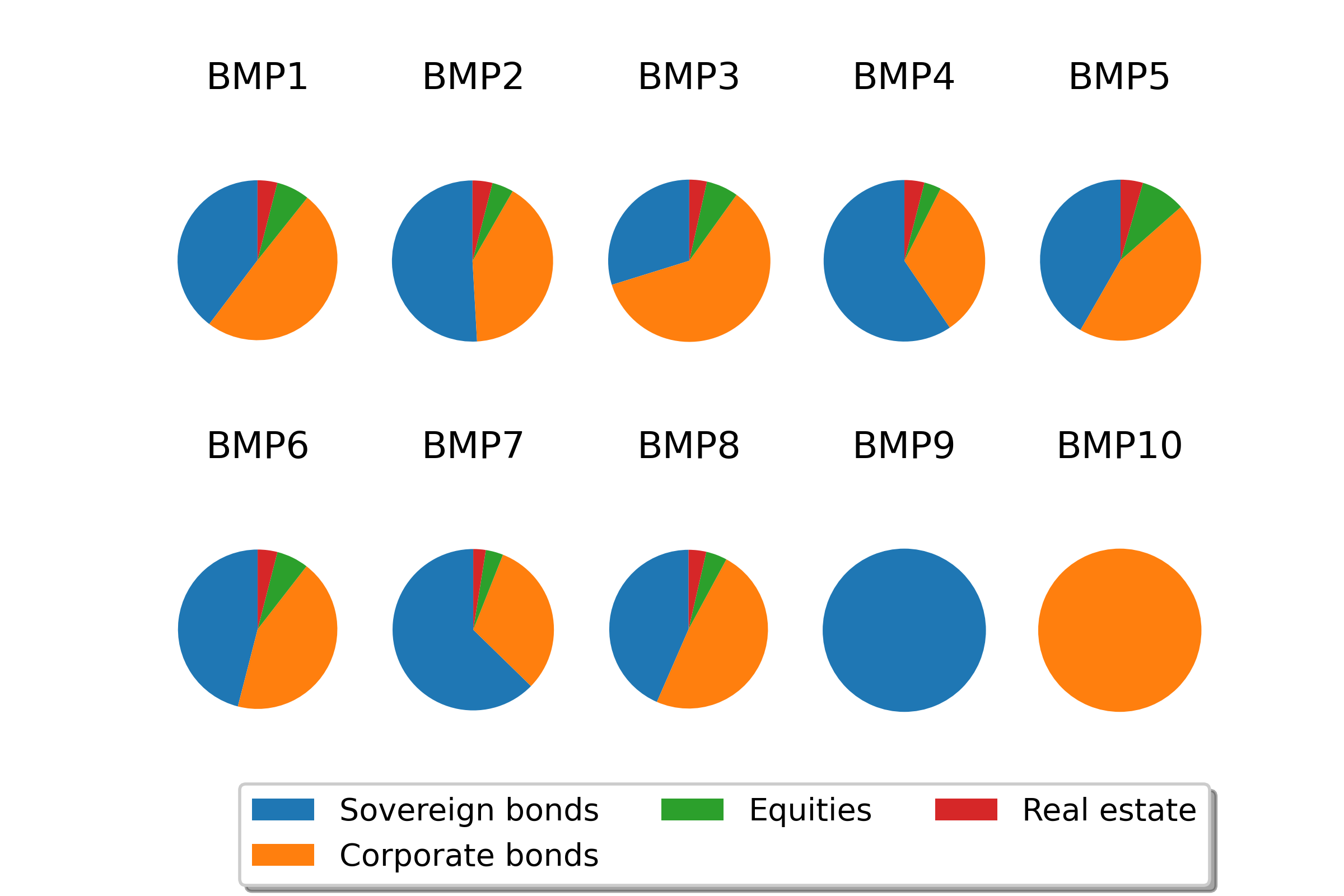}
\par\end{centering}
\caption{\label{fig:Composition-of-the-BMPs-1}Composition of the MCRCS asset-only
benchmark portfolios BMP1 - BMP10}
\end{figure}

Figure \ref{fig:Composition-of-the-BMPs-1} presents the asset-type
composition of the asset-only benchmark portfolios. 

All asset portfolios mainly consist of fixed income securities (86\%
to 94\%) as this forms the main investment focus of insurance companies.
However, there are significant differences both in ratings, durations
and also in the weighting between sovereign and corporate exposures.

The two liability profiles are assumed to be zero-bond based, so they
represent the liabilities of non-life insurance companies and differ
in their durations (13.1 years vs. 4.6 years).

Annually, EIOPA publishes a detailed article of the MCRCS exercise.
It provides an anonymized comparison of the risk charges of the different
insurance companies' market risk models by portfolios, instruments
and some additional analysis, e.g. dependencies of the risk factors.
The study for year-end 2019 can be found on the EIOPA homepage, see
\citet{EIOPA_MCRCS_Ergebnis}. We will use the results of this study
for comparison in Section \ref{sec:Comparison-of-GAN}.

\subsection{\label{subsec:Generative-adversarial-networks}Generative adversarial
networks}

Generative adversarial networks, called GANs, are an architecture
consisting of two neural networks which are interacting with each
other. In 2014, GANs were introduced by \citet{goodfellow2014pouget}
and have gained a lot of attention afterwards because of their promising
results especially in image generation. A good introduction to GANs
can be found in \citet{goodfellow2014pouget}, \citet{goodfellow2016nips}
and \citet{chollet2018deep}. According to \citet{motwani2020novel}
and \citet{li2020regularization}, GANs are one of the dominant methods
for the generation of realistic and diverse examples in the domains
of computer vision, image generation, image style transfer, text-to-image-translations,
time-series synthesis, natural language processing, etc.

Other popular methods for the generation of data based on empirical
observations are variational autoencoders and fully visible belief
networks, see \citet[p. 14]{goodfellow2016nips}.

Technically, a GAN consists of two neural networks, named \emph{generator}
and \emph{discriminator}. The discriminator network is trained to
distinguish real data points from ``fake'' data points and assigns
every given data point a probability of this data point being real.
The input to the generator network is random noise stemming from a
so called \emph{latent space}. The generator is trained to produce
data points that look like real data points and would be classified
by the discriminator as being real with a high probability. Figure
\ref{fig:Architecture-of-GAN} illustrates the general architecture
of a GANs training procedure:

\begin{figure}[H]
\begin{centering}
\includegraphics[width=8cm]{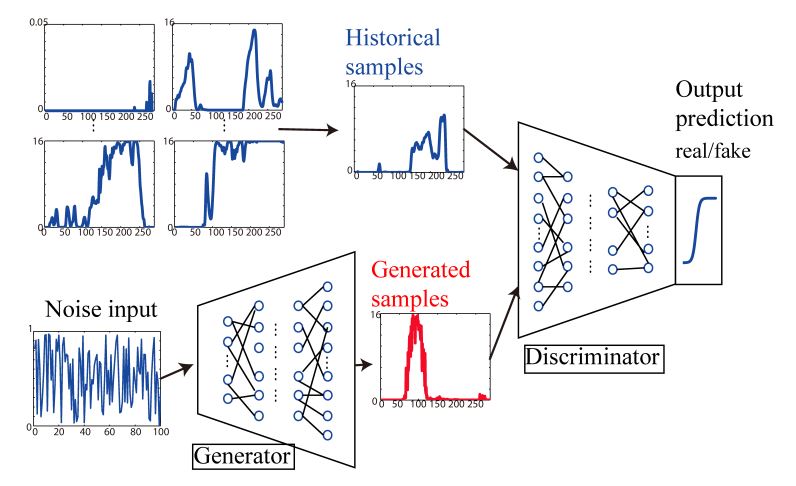}
\par\end{centering}
\caption{\label{fig:Architecture-of-GAN}Architecture of GANs training procedure,
according to \citet[p. 2]{chen2018bayesian}}
\end{figure}

Formally, we can define GANs as follows, see \citet[Chapter 3]{goodfellow2014pouget}
and \citet[Chapter 4]{wiese2020quant}. For this purpose, let $(\Omega,\mathcal{F},\ensuremath{\mathbb{P})}$
be a probability space and $N_{X},N_{Z}\in\mathbb{N}$. Furthermore,
assume that $X$ and $Z$ are $\R^{N_{X}}-$ and $\R^{N_{Z}}-$valued
random variables, respectively. The random variable $Z$ represents
the latent random noise variable and $X$ the targeted random variable.
$(\R^{N_{Z}},\mathfrak{\mathcal{B}(\R^{\mathrm{\mathit{N_{Z}}}}))}$
is called the \emph{latent space}. Usually, $N_{Z}>N_{X}$ is chosen,
see \citet[p. 18]{goodfellow2016nips}. The goal of the GAN is to
train a generator network, such that the generator mapping of the
random variable $Z$ to $\R^{N_{X}}$ has the same distribution as
the target variable $X$. Let's first define the generator formally:
\begin{defn}
Let $G_{\theta_{G}}:\R^{N_{Z}}\to\R^{N_{X}}$ be a neural network
with parameter space $\Theta_{G}$ and $\theta_{G}\in\Theta_{G}$.
The random variable $\widetilde{X}=G_{\theta_{G}}(Z)$ is called the
\emph{generated random variable}. The network $G_{\theta_{G}}$ is
called the \emph{generator.}
\end{defn}

The counterpart of the generator in this game is the discriminator
which assigns to each generated or real data point $x\in\R^{N_{X}}$
a probability of being a realization of the target distribution.
\begin{defn}
A neural network with $D_{\theta_{D}}:\R^{N_{X}}\to[0,1]$ with parameter
space $\Theta_{D}$ and $\theta_{D}\in\Theta_{D}$ is called a \emph{discriminator}.
\end{defn}

Given these two neural networks, we can now define a GAN as in \citet[Chapter 3]{goodfellow2016nips}:
\begin{defn}
A \emph{GAN (generative adversarial network) }is a network consisting
of a discriminator $D_{\theta_{D}}$ and a generator $G_{\theta_{G}}$.
The parameters $\theta_{D}\in\Theta_{D}$ and $\theta_{G}\in\Theta_{G}$
of both networks are trained to optimize the value function 
\[
V(G_{\theta_{G}},D_{\theta_{D}})=\mathbf{E}[\log(D_{\theta_{D}}(X))]+\mathbf{E}[\log(1-D_{\theta_{D}}(G_{\theta_{G}}(Z))]
\]
with $X$ the targeted random variable and $Z$ the latent random
noise variable.

$Z$ can be sampled from the latent space with a any chosen distribution,
however, one usually uses normal distribution for $Z$, see \citet[Chapter 8.5.2]{chollet2018deep}.
The value function $V$ has been defined in \citet{goodfellow2016nips}.
Where does it come from? The discriminator $D_{\theta_{D}}$ has to
distinguish between real and fake samples, i.e. solve a binary classification
problem, see \citet[Chapter 3.2]{goodfellow2016nips}. The discrete
version of the value function $V$ corresponds to the binary cross
entropy loss for the discriminator\foreignlanguage{english}{ which
}is usually used for binary classification issues solved by neural
networks. A definition of the binary cross entropy loss can be found
e.g. in \citet[Section 3]{ho2019real}\foreignlanguage{english}{\emph{.}}
\end{defn}

The optimization (the \emph{GAN objective}) is given by 
\[
\min_{\theta_{G}\in\Theta_{G}}\max_{\theta_{D}\in\Theta_{D}}V(G_{\theta_{G}},D_{\theta_{D}}).
\]
That means that the discriminator is optimized to distinguish real
samples from generated samples whereas the generator tries to `fool'
the discriminator by generating such good samples that the discriminator
is not able to distinguish them from real ones. A detailed derivation
of the optimization of the objective and the modifications that can
be used in GAN training are found in \citet[p. 22]{goodfellow2016nips}
and \citet[p. 9]{wiese2020quant}.

In the inner loop, $V(G_{\theta_{G}},D_{\theta_{D}})$ takes its maximum
value if the discriminator correctly assigns a value of 1 for all
``real'' data points and a value of 0 to all generated data points.
The parameters of the discriminator network $\theta_{D}$ are optimized
to fulfill this task. In the outer loop, the generator tries to fool
the discriminator and its parameters $\theta_{G}$ are optimized to
maximize $D_{\theta_{D}}(G_{\theta_{G}}(Z))$ meaning that the discriminator
shall assign a high probability of being real to the ``fake'' data
points. To achieve optimization of the value function, in practice,
the training alternates between $k$ steps of optimizing $D_{\theta_{D}}$
and one step of optimizing $G_{\theta_{G}}$. $k$ is one of the hyperparameters
of the GAN. The starting point of the parameters of the neural networks
$D_{\theta_{D}}$ and $G_{\theta_{G}}$ is given by random initialization.

An algorithm for the GAN training can be found in \citet[Chapter 4, Algorithm 1]{goodfellow2014pouget}.
In every iteration of the training process the neural networks are
trained in turns, while the parameters of the other network are fixed.
We provide here a short version of this algorithm:

\selectlanguage{ngerman}%
\begin{algorithm}[H]
\selectlanguage{british}%
The\foreignlanguage{ngerman}{ }discriminator\foreignlanguage{ngerman}{
}is\foreignlanguage{ngerman}{ trained $k$ times more often }than\foreignlanguage{ngerman}{
}the\foreignlanguage{ngerman}{ }generator\foreignlanguage{ngerman}{,
the dimension of the latent space $Z$ is $N^{Z}$, $M\in\N$ is the
batch size. All are hyperparameters of the GAN.}

\selectlanguage{ngerman}%
The learning rates of the SGD algorithm are $\gamma_{D}$ and $\gamma_{G}$.
\begin{itemize}
\item Initialize parameters $\boldsymbol{w}$ for discriminator and $\boldsymbol{\theta}$
for generator.
\item For each optimization step of the SGD:
\begin{itemize}
\item for $k$ steps do
\begin{itemize}
\item Randomly draw sample batch $\{\boldsymbol{x}_{1},...,\boldsymbol{x}_{M}\}$
of size $M$ from data generating distribution
\item Randomly sample $\{\boldsymbol{z}_{1},...,\boldsymbol{z}_{M}\}$ independent
realizations of random variable $Z$
\item Update the parameters $\boldsymbol{w}$ of the discriminator (with
fixed parameters $\boldsymbol{\theta}$ of the generator):
\[
\boldsymbol{w}_{neu}=\boldsymbol{w}_{alt}+\gamma_{D}\nabla_{\boldsymbol{w}}\frac{1}{M}\sum_{i=1}^{M}\left[\log(D_{\boldsymbol{w}}(\boldsymbol{x}_{i})+\log\left(1-D_{\boldsymbol{w}}(\boldsymbol{G}_{\boldsymbol{\theta}}(\boldsymbol{z}_{i})\right)\right].
\]
\end{itemize}
\item Randomly sample $\{\boldsymbol{z}_{1},...,\boldsymbol{z}_{M}\}$ independent
realizations of random variable $Z$
\item Update the parameters $\boldsymbol{\theta}$ of the generator (with
fixed parameters $\boldsymbol{\boldsymbol{w}}$ of the discriminator)
\[
\boldsymbol{\theta}_{neu}=\boldsymbol{\theta}_{alt}-\gamma_{G}\nabla_{\boldsymbol{\theta}}\frac{1}{M}\sum_{i=1}^{M}\left[\log\left(1-D_{\boldsymbol{w}}(\boldsymbol{G}_{\boldsymbol{\theta}}(\boldsymbol{z}_{i})\right)\right].
\]
\end{itemize}
\end{itemize}
\caption{\label{alg:Algorithm-for-GAN}Algorithm for GAN training with SGD
(stochastic gradient descent) as optimizer, see\foreignlanguage{british}{
\citet[Chapter 4, Algorithm 1]{goodfellow2014pouget}}}
\end{algorithm}

\selectlanguage{british}%
Despite their success, GANs remain difficult to train as stated e.g.
in \citet{motwani2020novel}. The largest issue with GANs according
to \citet[p. 34]{goodfellow2016nips} is the non-convergence often
observed in practice. This is due a GAN not being a normal optimization
task but a dynamic system seeking for an equilibrium between two forces,
see \citet[Chapter 8.5.2]{chollet2018deep} and \citet[p. 2]{salimans2016improved}.
Each time any parameter of one component, either discriminator or
the generator, is modified, it results in the instability of this
dynamic system, see \citet[p. 2]{motwani2020novel}. \citet[p. 25]{mazumdar2020gradient}
provide a mathematical analysis of this non-convergence issue and
state that ``first, the equilibrium {[}of the GAN{]} that is sought
is generally a saddle point and second, the dynamics of GANs are complex
enough to admit limit cycles``.

Therefore, the model architecture and the hyperparameter have to be
chosen carefully. Unfortunately, at the moment, there is no way to
tell which hyperparameters and which architecture will perform best
in the training, see \citet[p. 2]{motwani2020novel}. Therefore, some
kind of validation has to be performed to control convergence of the
GAN. One possible validation measure is the Wasserstein distance,
see Section \ref{subsec:Implementation-of-a}.\\

\begin{figure}[H]
\begin{centering}
\includegraphics[width=12cm]{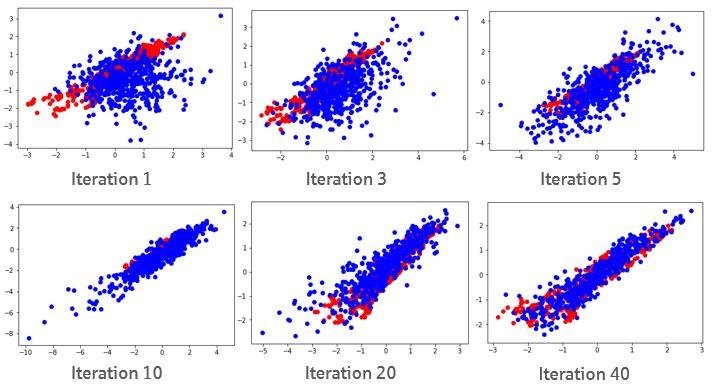}
\par\end{centering}
\caption{\label{fig:Scatterplots-by-iteration}Scatterplots of results of a
GAN for a two-dimensional data distribution at different training
iterations, red = empirical data, blue = generated data}
\end{figure}

The evolution of the quality of the output during training can be
visualized in case of a two-dimensional data distribution in Figure
\ref{fig:Scatterplots-by-iteration}. The red data here represents
the empirical data to be learned whereas the blue data is generated
at the current stage of training of the generator. For the illustration,
we here use the same amount of red and blue dots in each figure. One
can clearly see that the generated data matches the empirical data
more closely when more training iterations of the GAN have taken place.

\section{\label{sec:GAN-evaluation-measures-1}Methodology and data}

\subsection{Workflow of a GAN-based internal model}

The strength of GANs is especially what ESGs should be good at - producing
samples of an unknown distribution based on empirical examples of
that distribution. Therefore, we will apply a GAN as an ESG.

This is a different task from re-sampling, as explained e.g. in \citet[p. 3-7]{yu2002resampling},
where the task is to use subsets or bootstrapping from the empirical
data to derive data sets with the same statistical properties as the
empirical data. As we only have about 20 years of financial market
history and need to evaluate an 0.5-percentile, i.e. an one in 200
years event, we need to produce financial data that has not yet occurred
in the financial market but could have. Therefore, re-sampling is
not an option. The dependency modelling in a GAN-based ESG is similar
to using an empirical copula for the risk factors in a classical ESG
based on Monte-Carlo technique.

\begin{figure}[H]
\begin{centering}
\includegraphics[width=15cm]{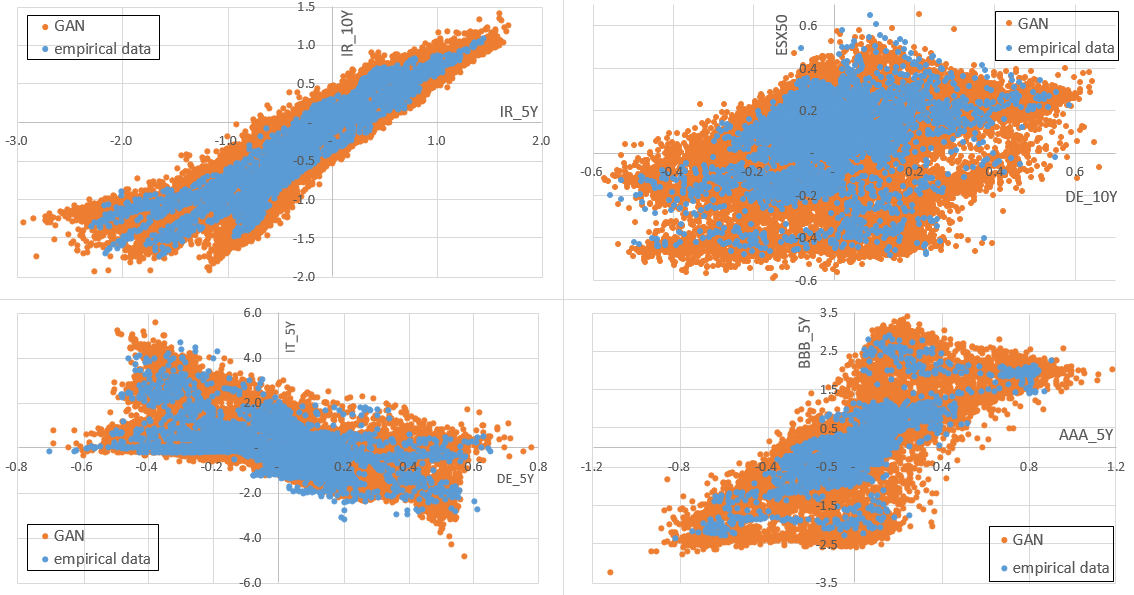}
\par\end{centering}
\caption{\label{fig:Scatterplots-of-four}Scatterplots of four different risk
factor pairs, empirical vs. generated data by a GAN}
\end{figure}

Figure \ref{fig:Scatterplots-of-four} shows that a GAN in contrast
to resampling really generates new scenarios. In this figure, we show
scatterplots of four different risk factor pairs: 5-year interest
rates vs. 10-year interest rates, Eurostoxx50 vs. German government
bond spreads, Italian vs. German government bond spreads and AAA corporate
credit spreads vs. BBB corporate credit spreads. Details to the data
can be found in Section \ref{subsec:Data-selection} and \ref{subsec:Data-preparation}.

The orange dots in the figure represent the 50,000 scenarios generated
by the GAN for these risk factors. The blue dots are the 4330 empirical
data points used for GAN training. The structure of the orange dots
mimics the structure of the blue dots, but also generates new results
that are not found within the blue dots. In resampling, the data would
not go beyond the boundaries of the blue dots.

\begin{figure}[H]
\begin{centering}
\includegraphics[width=10cm]{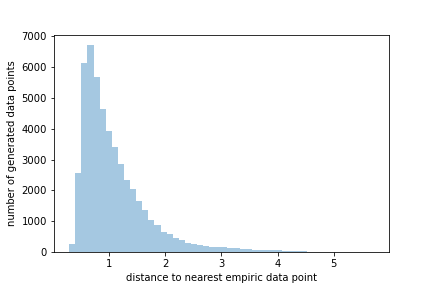}
\par\end{centering}
\caption{\label{fig:Histogram-of-minimum-distances}Histogram of minimum distances
of the generated data points to the empiric data points}
\end{figure}

For all 50,000 generated 46-dimensional scenario data points, we calculated
the euclidian distance to the nearest empirical data point in the
46-dimensional space. This distance is always above $0$ and varies
between 0.3 and 5.7. Figure \ref{fig:Histogram-of-minimum-distances}
shows the histogramm of those 50,000 minimum distances. In re-sampling,
the distance between generated and empiric data points always equals
0. This illustrates that the GAN really generates new scenarios.

As shown in \citet{chen2018bayesian}, a GAN can be used to create
new and distinct scenarios that capture the intrinsic features of
the historical data. \citet[p. 3]{fu2019time} already noted in their
paper that a GAN, as a non-parametric method, can be applied to learn
the correlation and volatility structures of the historical time series
data and produce unlimited real-like samples that have the same characteristics
as the empirical observed time-series. \citet[Chapter 5]{fu2019time}
tested this with two stocks and calculated a 1-day VaR.

In our work here, we demonstrate how to expand this to a whole internal
model - with enough risk factors to model the full band-width of investments
for an insurance company and for a one year time horizon as required
in Solvency 2.

\begin{figure}[H]
\begin{centering}
\includegraphics[width=10cm]{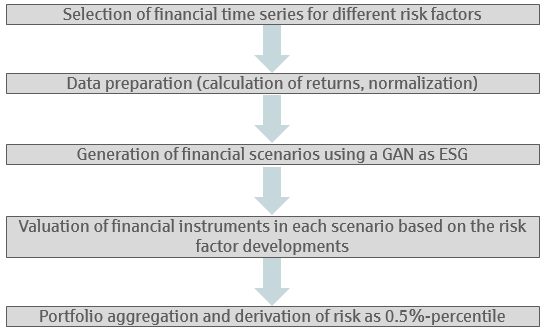}
\par\end{centering}
\caption{\label{fig:Workflow_internal_model}Workflow of a GAN-based internal
model}
\end{figure}

Diagram \ref{fig:Workflow_internal_model} illustrates schematically
the workflow of a GAN-based internal model. It follows the process
of a classical internal model, see \citet[p. 177 cont.]{bennemann2011handbuch}
and \citet[p. 82]{grundl2019solvency}, except that the ESG step is
replaced by a GAN instead of a Monte-Carlo simulation. Details to
the steps are provided in the next Sections and are similar to the
steps taken in \citet[Chapter 6]{wiese2020quant}.

\subsection{\label{subsec:Data-selection}Data selection}

For the purpose of this article, we choose to model the risk charge
for the ten MCRCS asset-only benchmark portfolios, the two liability-only
portfolios and for the ten combined asset-liability portfolios. We
are especially interested in the ability of the GAN to generate the
joint movement of the risk factors, and not the distribution of single
instruments. Therefore, we want to simulate only instruments and risk
factors that are needed in the benchmark portfolios.

Out of the 104 instruments in the MCRCS study, only 71 are included
in the benchmark portfolios; the 33 single instruments are left out
in our model. To model those instruments, we have to select financial
time series for the relevant risk factors (i.e. equity index, interest
rate buckets) as most of the instruments itself are not traded and
therefore have no time series. We found 46 risk factors to be sufficient
to evaluate those instruments because several instruments depend on
the same risk factor.

All relevant financial data will be derived from Bloomberg. An aggregated
view of these risk factors together with the Bloomberg sources (``ticker'')
can be found in Appendix \ref{sec:Appendix-2-Ticker}. A mapping table
indicating for each of the 71 instruments which risk factors are used
for calculation can be found in Appendix \ref{sec:Appendix-2-Instruments}.
In the same appendix, comments on any approximations used are found,
too.

EUR swap rates and EUR corporate yields are available on a daily basis
in Bloomberg since 25.02.2002 resp. 28.03.2002. Since the study is
conducted at year-end 2019, we take the datapool from end of March
2002 until Dec. 2019 as the basis of our GAN training. Therefore,
We use 4588 daily observations which covers almost 18 years of 46
time series to train the GAN.

\subsection{\label{subsec:Data-preparation}Data preparation}

For solvency 2, we need to model the market risk with a time horizon
of one year, but we have daily observations. According to \citet[p. 1]{yoon2019time}
this temporal setting poses a particular challenge to generative modeling.
One solution is to use the daily data to train the GAN model and then
use some autocorrelation function to generate an annual time series
based on daily returns, see \citet{fu2019time} and \citet[Chapter 34]{deutsch2004derivate}.
However, an autoregressive model can only be used if strong assumptions
about the underlying processes are made.

Another solution is to use overlapping rolling windows of annual returns
on a daily basis to train the model. We decide to calculate the returns
in rolling annual time windows for all available days. This method
is e.g. used in \citet[p. 16]{wiese2020quant} and also in \citet[p. 17]{EIOPA_MCRCS_Ergebnis}.
\citet{deutsch2004derivate} describes the procedure in Section 32.3.
He explains the drawback that the generated annual returns have a
high autocorrelation as they only differ in one daily return. However,
we will make the assumption that each of these one-year returns is
just one possible scenario how the risk factors can shift within one
year even if this implies the input data is not independent anymore.
This however is similar to classical ESGs where for calibration rolling
windows are often used as there is not enough data available to use
disjoint annual return data.

The second decision to be made is how to calculate the returns. Returns
can be calculated either as a simple difference of the two time points
in question, one can calculate relative returns, log-returns are often
used or one can use other transformations. \citet{EIOPA_MCRCS_Ergebnis}
in its analyses uses simple differences for interest rates and credit
spreads and relative returns for equities, real estate and foreign
exchange. The reasoning behind this is that in the low interest rate
environment, relative returns do not make sense in many cases. Therefore,
we will stick with this scheme.

To calculate rolling one-year returns, we make the assumption that
one year has 258 trading days. This is obtained by dividing the number
of daily data we have for the risk factors (T = 4588 observations)
by the number of years from which the data originate (17.8 years).

That means for each risk factor with a value of $s_{t}$ at time $t\in{1,...,T-258}$
with $t$ being the number of days in the data set, the rolling annual-return
is calculated as 
\[
r_{t}=\text{\ensuremath{\frac{s_{t+258}}{s_{t}}}}-1,\quad t=1,...,T-258.
\]
For interest rates and credit spreads, however, we calculate absolute
rolling returns as 
\[
r_{t}=s_{t+258}-s_{t},\quad t=1,...,T-258.
\]
Therefore, the training data for the GAN comprises $T-258=4330$ observations
of annual returns for each of the 46 risk factors.

A neural network does not work properly when the input data is at
different scales, see \citet[Chapter 3.6.2]{chollet2018deep}. Therefore,
we will normalize the return data $r_{t}$ by risk factor (dividing
by standard deviation, adjusting by the mean). As with GANs, there
is no split of the data into a training/test/validation data sets,
we can use all the data for training, see \citet[Chapter 3]{goodfellow2016nips}.

\subsection{\label{subsec:Implementation-of-a}Implementation of a GAN-based
ESG}

We implement the GAN in the programming language Python which has
a lot of packages that are useful in data science contexts. For pre-
and post-processing we use the packages pandas, numpy, scipy and matplotlib.
Our GAN implementation itself is based on the package keras as described
in \citet[Chapter 8.5]{chollet2018deep}.

It runs in a cloud with 144 virtual CPUs and 48 GiB RAM on a 3rd generation
Intel Xeon processor. The training time for this GAN is about half
an hour; the generation of the 50,000 scenarios as output takes less
than one minute. We used the following configuration for the GAN-based
ESG:

\begin{itemize}
\item $4$ layers for discriminator and generator
\item $400$ neurons per layer in the discriminator and $200$ in the generator
\item $k=10$ training iterations for the generator in each discriminator
training
\item Batch size is $M=200$
\item Dimension of the latent space is $200$, distribution of $Z$ is multivariate
normal with mean = $0$ and std = $0.02$
\item Initialization of generator and discriminator using multivariate normal
distribution with mean = $0$ and std = $0.02$
\item We use LeakyReLu as activation functions except for the output layers
which use Sigmoid (for discriminator) and linear (for generator) activation
functions. We use the Adam optimizer and the regulation technique
batch normalization after each of the hidden layers in the network.
The loss function is binary crossentropy.
\end{itemize}
LeakyRelu hereby is an alternative to the popular activation function
Relu and is defined as 
\[
g(x)=\begin{cases}
x & x>0\\
\alpha x & x<0
\end{cases}
\]

In our implementation, we set $\alpha=0.2$.

The adam optimizer is used here instead of the SGD, see Algorithmus
\ref{alg:Algorithm-for-GAN}. The adam optimizer uses the following
algorithm to update the weights to $w_{k+1}$ given $w_{k}$ in the
generator and discriminator networks:

\selectlanguage{ngerman}%
\begin{align*}
\boldsymbol{v}_{k} & =\beta_{2}\boldsymbol{v}_{k-1}+(1-\beta_{2})\left(\boldsymbol{g}_{k}\astrosun\boldsymbol{g}_{k}\right),\quad k=1,2...\\
\boldsymbol{m}_{k} & =\beta_{1}\boldsymbol{m}_{k-1}+(1-\beta_{1})\boldsymbol{g}_{k},\quad k=1,2...\\
\boldsymbol{w}_{k+1} & =\boldsymbol{w}_{k}+\gamma_{k}\frac{\sqrt{1-\beta_{2}^{k+1}}}{1-\beta_{1}^{k+1}}\left(\boldsymbol{m}_{k}\astrosun\frac{1}{\sqrt{\boldsymbol{v}_{k}}+\delta}\right),\quad k=0,1,...
\end{align*}
with $\boldsymbol{v}_{0}$ und $\boldsymbol{m}_{0}$ being null vectors
and \emph{$\astrosun$} the\emph{ }Hadamard product. The parameters
used in our implementation are
\begin{align}
\gamma_{0}=\gamma_{2}=... & =0.0002\label{eq:adam}\\
\beta_{1} & =0.5\nonumber \\
\beta_{2} & =0.999\nonumber \\
\delta & =10^{-7}\nonumber 
\end{align}

\selectlanguage{british}%
Definitions and explanations for these functions and terms can be
found in e.g. \citet{viehmann2019variants} and \citet{chollet2018deep}.

According to \citet[Chapter 5.2]{goodfellow2016nips} ``it is not
clear how to quantitatively evaluate generative models. Models that
obtain good likelihood can generate bad samples, and models that generate
good samples can have poor likelihood. There is no clearly justified
way to quantitatively score samples''. According to \citet[p. 1]{theis2015note}
``generative models need to be evaluated directly with respect to
the application(s) they were intended for''.

In literature, where scenario generation for financial or similar
data is done by GANs or other generative models, there is no clear
favorite measure being used for validation of the results. Many papers
focus on visual inspection of histograms, scatterplots, etc. together
with the comparison of some basic statistics (mean, standard deviation,
skewness, kurtosis, percentiles, (auto)correlation). Examples for
this evaluation method are the following papers: \citet{franco2019generating},
\citet{henry2019generative}, \citet{lezmi2020improving}, \citet{fu2019time},
\citet{chen2018bayesian} and \citet{marti2020corrgan}.

As for internal models, the marginal distribution of the risk factors
are of great importance, we will use Wasserstein distance in this
paper to evaluate whether the generated data matches the empirical
data. The \emph{Wasserstein distance} is commonly used to calculate
the distance between two probability distribution functions, as mentioned
in \citet{borji2019pros} and \citet{wiese2020quant}. The definition
of the Wasserstein distance can be found e.g. in \citet[p. 5]{hallin2021multivariate}.
We use here the univariate Wasserstein distance with $p=1$ as defined
in \citet[eq. 1]{hallin2021multivariate}:
\begin{defn}
For univariate distributions $P$ and $Q$ with distribution functions
$F$ and $G$, the $p$-Wasserstein distance is given by the $L^{p}$-distance
\[
\mathcal{W_{\mathrm{p}}}(P,Q)=\bigg(\int_{0}^{1}\Big|F^{-1}(t)\text{\textminus}G^{-1}(t)\Big|^{p}dt\bigg)^{1/p}.
\]
\end{defn}

One could alternatively use other metrics to measure the distance
between two probability distribution functions, e.g. Kullback-Leibler
divergence. We decided for the Wasserstein distance as it is easily
interpretable, implementable and often used in the machine learning
context.

\begin{figure}[H]
\begin{centering}
\includegraphics[width=10cm]{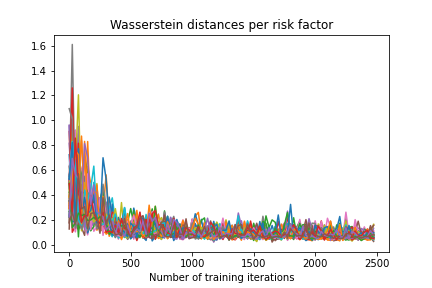}
\par\end{centering}
\caption{\label{fig:Development-of-measures_2500}Development of the Wasserstein
distances for all 46 risk factors during GAN training iterations}
\end{figure}

Figure \ref{fig:Development-of-measures_2500} shows the development
of the Wasserstein distances between the empirical distribution functions
of the training and the generated data for all 46 risk factors in
our GAN. We can see in this graph that the Wasserstein distance for
all risk factors decreases over the training iterations in this configuration.

To arrive at the GAN architecture mentioned above, we trained 25 different
GANs varying in the number of layers and the number of neurons per
layer. For each of the 25 configurations, we then calculated the maximum
of the Wasserstein distances over all the risk factors. We then compared
the minimum of this maximal Wasserstein distance over the training
iterations between the 25 configurations and choose the configuration
having the lowest value. Details and results of this procedure can
be found in Appendix \ref{sec:Appendix-3-Optimization}.

We now use the trained generator with this configuration to generate
50,000 financial scenarios for all risk factors.

\subsection{\label{subsec:Implementation}Valuation of financial instruments
and portfolio aggregation}

As mentioned before, we want to evaluate the benchmark portfolios
of MCRCS study which comprise 71 instruments. So, first, we have to
evaluate the instruments in each scenario applying the generated data
for the 46 risk factors. For this task, we use the following formulas
and methods.\\

\emph{1. Zero-coupon bond valuation}

All interest-and spread-related instruments are zero-coupon bonds
and can be valued with present value discounting for scenario $n=1,...,50000$
as 
\[
ZC(r_{\tau}^{0},\triangle r_{\tau}^{n},s_{\tau}^{0},\triangle s_{\tau}^{n},\tau)=\frac{1}{(1+r_{\tau}^{0}+\triangle r_{\tau}^{n}+s_{\tau}^{0}+\triangle s_{\tau}^{n})^{\tau}}
\]

where $ZC$ means the value of a zero-coupon bond with maturity $\tau$
with starting interest rate being $r_{\tau}^{0}$ and the shift in
scenario $n$ being $\triangle r_{\tau}^{n}$ and the starting spread
being $s_{\tau}^{0}$ and the shift in scenario $n$ of the spread
being $\triangle s_{\tau}^{n}$ . $\tau$ hereby equals the maturity
of the bond as found in Table \ref{tab:Mapping-of-instruments} in
Appendix \ref{sec:Appendix-2-Instruments}. The starting time for
MCRCS study is year-end 2019. For details on the valuation of zero
coupon bonds, we refer to \citet[Chapter 8.4]{albrecht2016investment}.

For the default and migration process of corporate bonds, we use the
rating migration matrix from \citet{SP_RatingTransistions}. This
is the S\&P average European 1-year corporate transition rate for
the years 1981-2018where we assume that high yield bonds are B-rated.
Since the credit spread can be interpreted as the probability of a
bond defaulting, we scale the downgrade probabilities in the scenarios
according to the percentage the generated spread is above the starting
spread. For sovereign bonds, we analogously assumed a default according
to the rating of the country, scaled using the same methodology as
for corporate bonds. The recovery rate is set at 45\%.\\

\emph{2. Equity and property instrument valuation}

For equities and property, the market value of the instruments in
the scenarios is scaled with the percental shift of the respective
risk factor to evaluate those instruments in scenario $n$.\\

\emph{3. Valuation of the liabilities}

The method used for discounting of the liabilities in Solvency 2 is
laid out in \citet{EIOPA_Zinskurve_TechnicalDocumentation}. The Solvency
2 framework assumes that liquidity in the interest rate market can
only be assumed up to a maturity of 20 years. In this period, a credit
risk adjustment of 10 basispoints is deducted before discounting.
Afterwards, the yield curve is extrapolated to a so-called \emph{ultimate
forward rate}, abbr. ``UFR'', which is the 1-year forward rate to
be valid at a maturity of 60 years. The UFR is updated on an annual
basis and at year-end 2019 was 3.9\%. The extrapolation method is
based on \citet{smith2001fitting} and its features are discussed
e.g. in \citet{viehmann2019variants} and \citet{lageraas2016issues}.
In our implementation, after generating the scenarios, we use this
extrapolation method to derive the risk-free yield curve and discount
the two liability benchmark portfolios accordingly.\\

\emph{4. Portfolio aggregation}

After calculating the profit\&loss of each of the relevant instruments
in each scenario, the portfolio aggregation is straightforward using
the weights given in the EIOPA MCRCS study, see \citet{EIOPA_MCRCS_Excel}.
The risk charge then is the 0.5\%-percentile of the 50.000 scenarios
for each portfolio.

\section{\label{sec:Comparison-of-GAN}Comparison of GAN results with the
results of the MCRCS study}

Now we can compare the results of our GAN-based model for both risk
factors and benchmarking portfolios with the risk derived from approved
internal models in Europe using the results of the MCRCS study. The
study for year-end 2019 can be found on EIOPA's homepage, see \citet{EIOPA_MCRCS_Ergebnis}.

The results on risk factor basis are analyzed based on the shocks
generated or implied by the ESGs in the study in Section \ref{subsec:Comparison-on-risk-factor}.
A shock hereby is defined in \citet[p. 11]{EIOPA_MCRCS_Ergebnis}
as
\begin{defn}
A \emph{shock} is the absolute change of a risk factor over a one-year
time horizon. Depending on the type of risk factor, the shocks can
either be two-sided (e.g. interest rates 'up/down') or one-sided (e.g.
credit spreads 'up'). This metric takes into account the undertakings\textquoteright{}
individual risk measure definitions and is based on the 0.5\% and
99.5\% quantiles for two-sided risk factors and the 99.5\% quantile
for one-sided risk factors, respectively.
\end{defn}

The main comparison between the results for the the benchmark portfolios
is based on the risk charge which is defined in \citet[Section 2]{EIOPA_MCRCS_Instructions}.
We analyze this in Section \ref{subsec:Comparison-on-portfolio}.
\begin{defn}
The\emph{ risk charge} is the ratio of the modeled Value at Risk (99.5\%,
one year horizon) and the provided market value of the portfolio.
\end{defn}

The results are presented in diagrams, showing the 10\%, 25\%, 75\%
and 90\%-percentile of all insurance companies participating in the
study. There are 21 participants and only insurances having at least
some exposure in this risk factor are shown. For our comparison here,
we want to compare our results with the whole bandwidth of internal
models in Europe. Therefore, we calculated an implied mean and standard
deviation based on the 10\%- and 90\%-percentile under the assumption
of a normal distribution of results. Based on this, we derived a theoretical
1\%- and 99\%-percentile to show as boxes for each maturity / sub-type
of the risk factors resp. portfolios. The given 10\%- and 90\%-percentile
are shown as frames inside the boxes.

We enrich those MCRCS results with a blue dot representing the shock
resp. risk charge for that risk factor or portfolio generated by our
GAN-based model.

After that, in Section \ref{subsec:Comparison-of-the-Covid}, we use
the results shown in a so-called excursus focusing on the market development
in the Covid-19 crisis to compare the results of the approved models
with our GAN-based model, too. The dependency structures of the internal
models are analyzed using joint quantile exceedance as a metric in
\citet[Chapter 5.2.6]{EIOPA_MCRCS_Ergebnis}. The comparison to the
GAN results can be found in Section \ref{subsec:Results_JQE}. The
Section concludes with an examination of the stability of the GAN
output.

\subsection{\label{subsec:Comparison-on-risk-factor}Comparison on risk-factor
level}

In this work, we will show the comparison of the five most important
risk factor categories (corporate and sovereign credit spread, equity,
interest rate up and down). Other risk factors show a similar behaviour.

\begin{figure}[H]
\begin{centering}
\includegraphics[width=15cm]{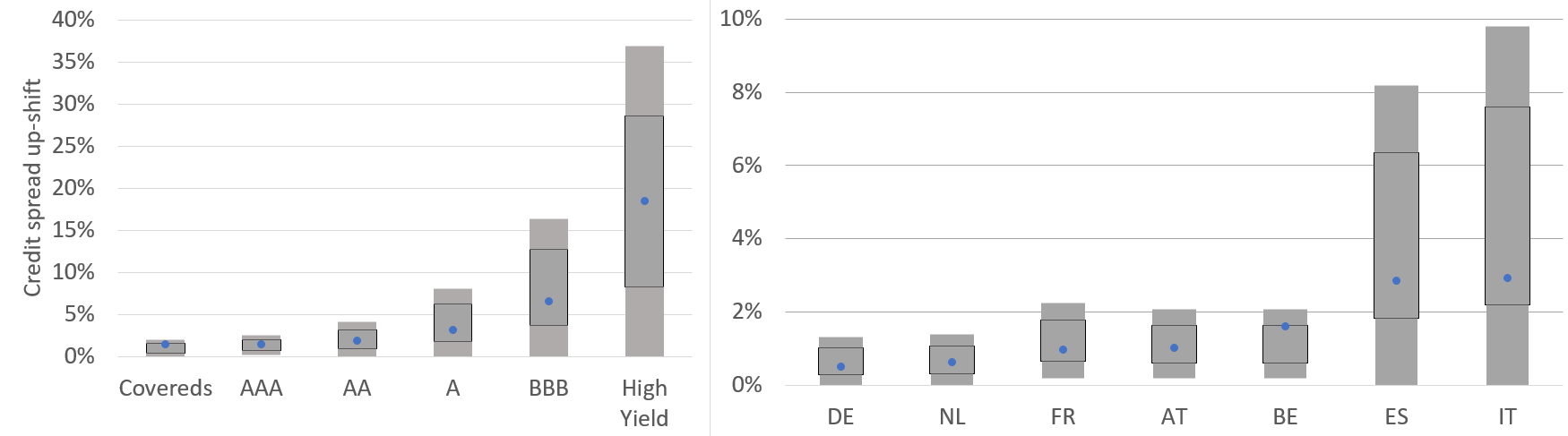}
\par\end{centering}
\caption{\label{fig:Comparison-of-CS}Comparison of the simulated shifts for
the corporate and sovereign credit spread risk factors, representation
based on own results (blue dots) and \citet[p. 25 and 26]{EIOPA_MCRCS_Ergebnis}
(gray boxes)}
\end{figure}

For corporate as well as sovereign credit spreads in Figure \ref{fig:Comparison-of-CS},
we see a very good alignment between the GAN-based model and the approved
internal models. The comparison of the Ireland sovereign spread has
been excluded from the comparison as only five participants submitted
results for this risk factor. The shock reported in \citet[p. 26]{EIOPA_MCRCS_Ergebnis}
varies between 1.1\% and 3.5\% which seems inconsistent to the sharp
increase of up to 8.3\% within 12 months that Ireland experienced
during financial crisis. Therefore, data quality here seems not to
be sufficient for a comparison.

\begin{figure}[H]
\begin{centering}
\includegraphics[width=7.5cm]{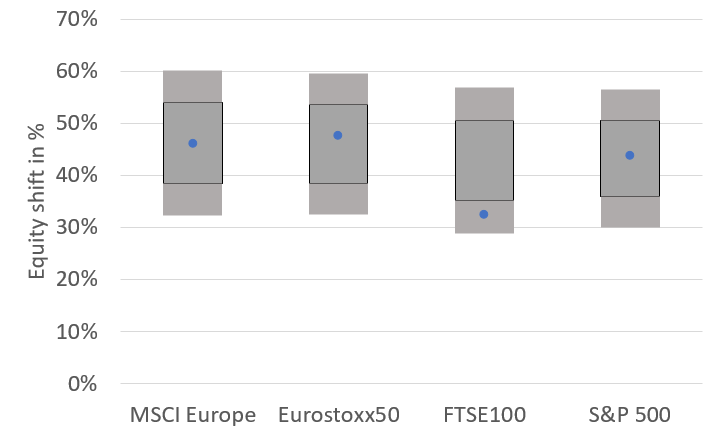}
\par\end{centering}
\caption{\label{fig:Comparison-of-EQ}Comparison of the simulated shifts for
equity risk factors, representation based on own results (blue dots)
and \citet[p. 27]{EIOPA_MCRCS_Ergebnis} (gray boxes)}
\end{figure}

On the equity side in Figure \ref{fig:Comparison-of-EQ}, the shifts
are also similar for most of the risk factors. For the FTSE100, the
GAN produces less severe shocks than most of the other models. This
behaviour, however, can actually be found in the training data as
the FTSE100 is less volatile than the other indices for the time frame
used in GAN training. So, the GAN here produces plausible results.

For interest rates, however, the picture is a bit more complex as
Figure \ref{fig:Comparison-of-IR} illustrates:

\begin{figure}[H]
\begin{centering}
\includegraphics[width=7.5cm]{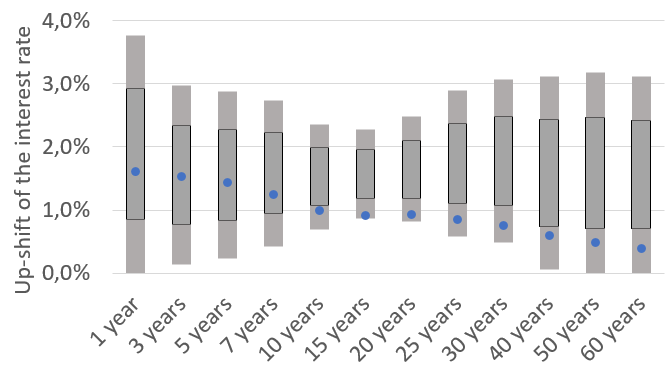}\includegraphics[width=7.5cm]{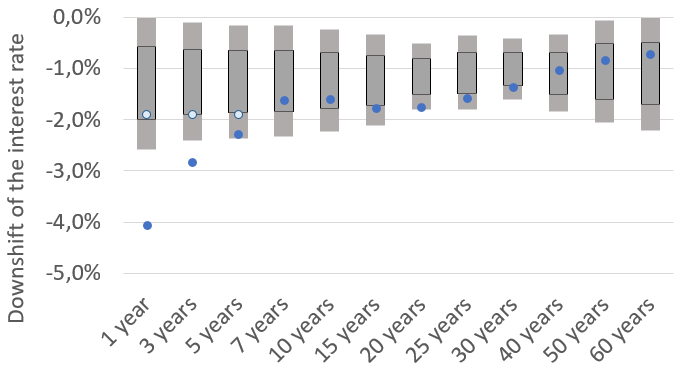}
\par\end{centering}
\caption{\label{fig:Comparison-of-IR}Comparison of the simulated shifts (left:
up, right: down) for the interest rate risk factors, representation
based on own results (blue dots) and \citet[p. 22]{EIOPA_MCRCS_Ergebnis}
(gray boxes)}
\end{figure}

The up-shifts generated by the GAN-based model are within the boxes
for all buckets. However, for longer maturities, the shifts tend to
be at the lower end of the boxes. This effect is due to the time span
of the data used for the training of the GAN where interest rates
are mostly decreasing.

For the down-shifts, we can observe the following behaviour: Short
term interest rates are below the boxes, whereas the middle and longer
term interest rates are inside the boxes. This can be explained by
the interest rate development: The time span used for training of
the GAN shows a sharp decrease of interest rates especially in the
short term whereas longer term interest rates behaved more stable.
This behaviour is mimicked by the GAN. In traditional ESGs, additionally
to the longer time span used for calibration, often expert judgement
by the insurers leads to a lower bound on how negative interest rates
can become. One of the most common arguments for a lower bound of
interest rates according to \citet[Chapter 4]{grasselli2019normality}
is the fact that instead of investing money with negative interest
rates, asset managers could also convert the money into cash and store
this. However, the conversion of large amounts of money into cash
poses a lot of issues and is therefore unrealistic. \citet{grasselli2019normality}
use this argument to derive a cash-related physical lower bound of
about -0.5\%. \citet[Chapter 2]{danthine2017interest} states that
the lower boundary for interest rates is not far below -0.75\% in
current environment. For illustration purposes, we introduced light
blue dots in Figure \ref{fig:Comparison-of-IR} where we limited the
downshift to -1.9\% in the GAN (as this is the value of the 10\%-percentile).
This, however, doesn't change the results on the portfolio level as
presented in Section \ref{subsec:Comparison-on-portfolio} significantly
(difference in VaR always below 0.1\%). If an insurance company wishes
to limit downside interest rate shifts, this would be a reasonable
approach.

\subsection{\label{subsec:Comparison-on-portfolio}Comparison on portfolio level}

First, we show the comparison of the risk charges for the ten asset-only
benchmark portfolio of MCRCS study:

\begin{figure}[H]
\begin{centering}
\includegraphics[width=12cm]{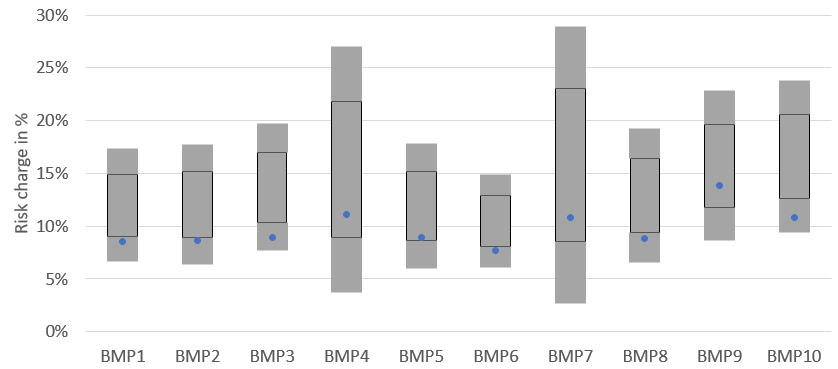}
\par\end{centering}
\caption{Comparison of the asset-only benchmark portfolios, representation
based on own results (blue dot) and \citet[p. 16]{EIOPA_MCRCS_Ergebnis}
(gray boxes)}
\end{figure}

The risk charge of the GAN-based model fits well to the risk charges
of the established models and always lays within the gray boxes. The
blue dot tends to be at the lower part of the boxes for the portfolios.
This is due to the fact that increasing interest rates form a main
risk. As we showed in Section \ref{subsec:Comparison-on-risk-factor},
the shocks for the GAN-based model for increasing interest rates are
at the lower part of the boxes for longer maturities, too. So this
behaviour can be explained.

The two liability-only portfolios differ by duration: L1 has a duration
of 13.1 years versus a 4.6 years duration of L2.

\begin{figure}[H]
\begin{centering}
\includegraphics[width=6cm]{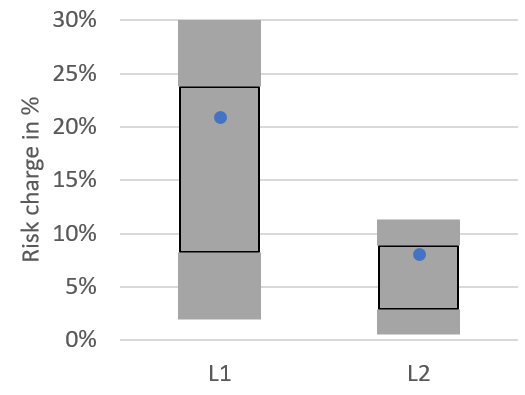}
\par\end{centering}
\caption{Comparison of the liabilities-only benchmark portfolios, representation
based on own results (blue dot) and \citet[p. 25]{EIOPA_MCRCS_Ergebnis}
(gray boxes)}
\end{figure}

For the liability-only portfolios, too, the risk charge of the GAN-based
model fits well to the risk charges of the established models and
always lays within the gray boxes. The blue dot tends to be at the
upper part of the boxes for the portfolios. The risk charge of the
liability-only portfolios is caused by scenarios with decreasing interest
rates. In the risk factor comparison of the interest rate down shock
in \ref{subsec:Comparison-on-risk-factor}, the interest rate down
shifts tend to be more severe for most maturities for the GAN-based
model than for the average of the other models. Therefore, it seems
plausible for the resulting portfolio risk to be at the upper part,
too.

\begin{figure}[H]
\begin{centering}
\includegraphics[width=7.5cm]{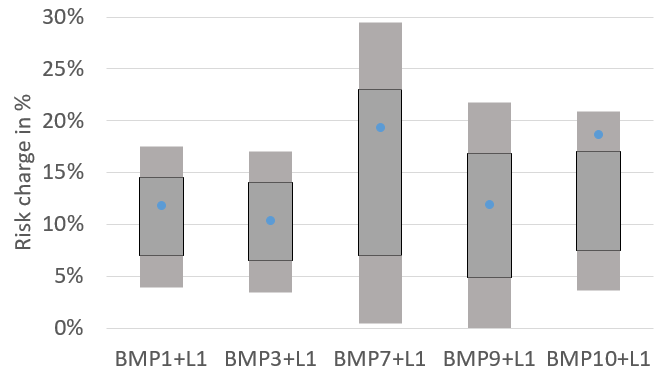}\includegraphics[width=7.5cm]{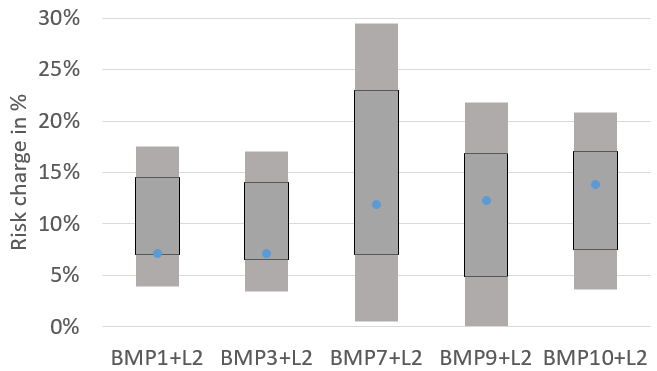}
\par\end{centering}
\caption{\label{fig:Comparison-of-BMPAL-1}Comparison of the combined asset-liability
benchmark portfolios, representation based on own results (blue dot)
and \citet[p. 14]{EIOPA_MCRCS_Ergebnis} (gray boxes)}
\end{figure}

Figure \ref{fig:Comparison-of-BMPAL-1} displays the risk charge for
each of the combined asset-liability benchmark portfolios which comprises
portfolios with the longer (left) and the shorter liability structure
(right). 

For the combined portfolios, too, the GAN-based model shows comparable
results to the established models.

\subsection{\label{subsec:Comparison-of-the-Covid}Comparison of the Covid-19
backtesting results}

In a so called excursus, \citet[p. 17-21]{EIOPA_MCRCS_Ergebnis},
the study examines whether the turmoil in the financial markets following
the Covid-19 crisis in spring 2020 is part of the generated scenarios
of the tested models. This could be seen as a kind of backtesting
exercise as this event was not part of the calibration / training
process for the ESGs. In the study, EIOPA calculates for each benchmark
portfolio $P$ the worst case one-year rolling return including the
Covid-19 crisis by 
\begin{equation}
WorstCase(P)=\min_{t\in\mathbb{T}}\Big(\frac{MarketValue(P)_{t}}{MarketValue(P)_{t-258}}-1\Big),\label{eq:worstCase}
\end{equation}
where
\[
\mathbb{T}=\left\{ \text{Last working day for each month in the period 31.01.2017 - 30.09.2020}\right\} .
\]
Let $F_{P,M}$ be the empirical distribution function under model
$M$ for the relative returns of portfolio $P$ with respect to the
$50,000$ scenarios relative to the current market value. By
\[
\alpha_{P,M}=F_{P,M}\big(WorstCase(P)\big)
\]
we denote the probability that a relative loss as least as severe
as the Covid-19 crisis occurs to portfolio $P$ under model $M$.

\citet[p. 21]{EIOPA_MCRCS_Ergebnis} states that ``the Covid-19 related
market impacts can certainly be seen as significant. From a general
perspective of internal market risk models, there is no evidence that
this could be interpreted as an event beyond the scope of application
of these models.'' This means that the worst-case losses in this
time period considered for every benchmark portfolio should be within
the loss distribution generated by the models and ideally $\alpha_{P,M}$
is above 0.5\%. The worst case for most portfolios occurs during Covid-19-related
market turmoil in the first half of 2020, see \citet[p. 19]{EIOPA_MCRCS_Ergebnis}.

Explicit comparison of the values $\alpha_{P,M}$ for the models is
only provided for the asset-only benchmark portfolio BMP1 and the
asset-liability benchmark portfolio BMP1+L1. As in Sections \ref{subsec:Comparison-on-risk-factor}
and \ref{subsec:Comparison-on-portfolio} above, the same meaning
of the boxes apply in the graph.

\begin{figure}[H]
\begin{centering}
\includegraphics[width=10cm]{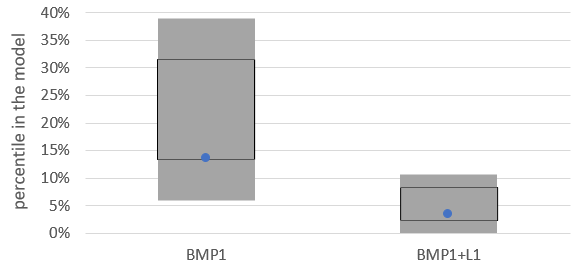}
\par\end{centering}
\caption{Comparison of the implied percentiles for the backtesting exercise,
representation based on own results (blue dots) and \citet[p. 19-20]{EIOPA_MCRCS_Ergebnis}
(gray boxes)}
\end{figure}

The results match with the results of Section \ref{subsec:Comparison-on-portfolio}.
One can state that the GAN-based model in this examination, too, behaves
similar to the other models. For the other benchmark portfolios, we
also calculate the implied percentiles of the WorstCase return in
spring 2020 and the corresponding value $\alpha_{P,GAN}$ for the
GAN-based model $\alpha_{P,GAN}$. In the Table \ref{tab:Worst-case-losses},
for portfolio BMP1+L1 a value of $\alpha_{P,GAN}=3.5\%$ means that
in $3.5\%$ of the scenarios the GAN-based model generated a return
more severe than $-58.2\%$ with $-58.2\%$ being the WorstCase return
of portfolio BMP1+L1 encountered during Covid-19 crisis.

As displayed in Table \ref{tab:Worst-case-losses}, for all portfolios
$\alpha_{P,GAN}$ for the GAN-based model is above 0.5\% which is
in line with the EIOPA expectations mentioned above.

\begin{table}[H]
\begin{centering}
\begin{tabular}{|>{\centering}p{3.5cm}|>{\centering}p{3.5cm}|>{\centering}p{3.5cm}|}
\hline 
Benchmark portfolio &
$WorstCase(P)$ &
$\alpha_{P,GAN}$\tabularnewline
\hline 
\hline 
BMP1 &
-2.8\% &
13.5\%\tabularnewline
\hline 
BMP2 &
-2.5\% &
19.2\%\tabularnewline
\hline 
BMP3 &
-2.8\% &
15.6\%\tabularnewline
\hline 
BMP4 &
-2.7\% &
19.6\%\tabularnewline
\hline 
BMP5 &
-2.4\% &
13.3\%\tabularnewline
\hline 
BMP6 &
-2.8\% &
10.9\%\tabularnewline
\hline 
BMP7 &
-7.8\% &
2.3\%\tabularnewline
\hline 
BMP8 &
-2.7\% &
16.2\%\tabularnewline
\hline 
BMP9 &
-3.7\% &
21.6\%\tabularnewline
\hline 
BMP10 &
-6.1\% &
5.9\%\tabularnewline
\hline 
L1 &
-15.2\% &
5.4\%\tabularnewline
\hline 
L2 &
-4.3\% &
15.8\%\tabularnewline
\hline 
BMP1+L1 &
-58.2\% &
3.5\%\tabularnewline
\hline 
BMP3+L1 &
-64.5\% &
0.9\%\tabularnewline
\hline 
BMP7+L1 &
-67.1\% &
6.8\%\tabularnewline
\hline 
BMP9+L1 &
-41.3\% &
7.1\%\tabularnewline
\hline 
BMP10+L1 &
-89.9\% &
3.4\%\tabularnewline
\hline 
BMP1+L2 &
-30.9\% &
3.8\%\tabularnewline
\hline 
BMP3+L2 &
-34.8\% &
2.5\%\tabularnewline
\hline 
BMP7+L2 &
-53.4\% &
3.9\%\tabularnewline
\hline 
BMP9+L2 &
-26.8\% &
13.6\%\tabularnewline
\hline 
BMP10+L2 &
-65.6\% &
3.6\%\tabularnewline
\hline 
\end{tabular}
\par\end{centering}
\caption{\label{tab:Worst-case-losses}Worst case losses of all benchmark portfolios
and implied percentiles of the GAN-based model}
\end{table}

\subsection{\label{subsec:Results_JQE}Comparison of joint quantile exceedance
results}

The market-risk dependency structures in the models are examined in
\citet[Chapter 5.2.6]{EIOPA_MCRCS_Ergebnis} on a risk factor basis.
Results are only presented for the comparison of joint quantile exceedance
which is defined as follows:
\begin{defn}
The bivariate Joint Quantile Exceedance probability (JQE) is the joint
probability that both risk factors will simultaneously surpass the
same quantile.
\end{defn}

For the comparison a percentile of 80\% is used. This is a compromise
to have enough data to examine, but also to focus on the tail of the
distribution. For independent risk factors, the joint quantile exceedance
therefore equals $JQE=20\%\cdot20\%=4\%$. If the risk factors have
a correlation of 1, the JQE equals $20\%$; for a correlation of -1,
JQE is $0\%$. In the study, \citet[p. 34]{EIOPA_MCRCS_Ergebnis},
the matrix of JQEs is presented as boxplots for all pairs of 7 selected
risk factors. Please note that the joint quantile exceedance of the
risk factor with itself is not shown.

We show here the results of the comparison for one credit spread,
one equity and one interest rate risk factor. Other comparisons follow
a similar pattern.

\begin{figure}[H]
\begin{centering}
\includegraphics[width=11cm]{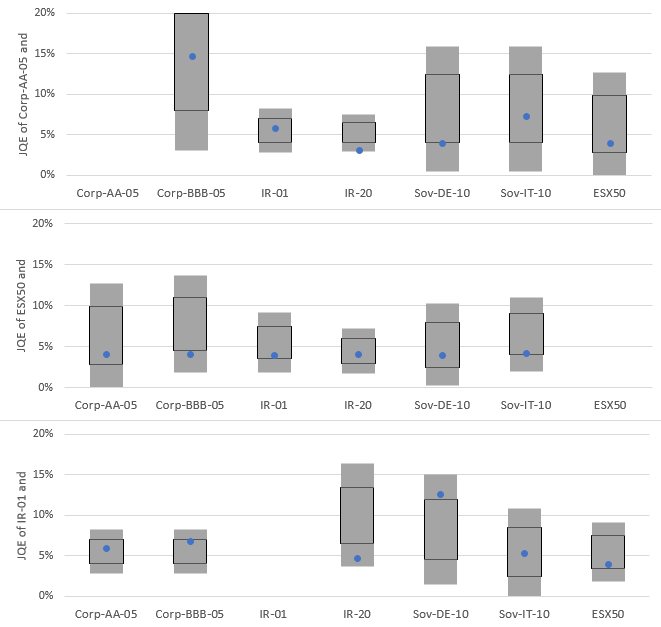}
\par\end{centering}
\caption{Comparison of the joint quantile exceedance for selected risk factors,
representation based on own results (blue dots) and \citet[p. 35]{EIOPA_MCRCS_Ergebnis}
(gray boxes)}
\end{figure}

For the pairwise JQE results, the GAN-based model always lays within
the boxes of the internal models of the MCRCS study. Therefore, the
dependency structure generated by the GAN-based ESG seems to resemble
the dependency structures used in internal market risk models in Europe.

\subsection{\label{subsec:Stability-of-GAN}Stability of GAN results}

One important question for an internal model is how stable the results
of the GAN-based ESG are. Since the GAN is initialized with random
parameters and the sampling for the batches in training is random
as well as the generation of the random variables in latent space,
we want to test whether the results in the previous sections are stable
for different GAN runs.

We trained four different GANs with the same architecture (but with
different random initialization) and used each of those four trained
generators to generate$50,000$ scenarios 5 times for each of the
46 risk factors, leading to 20 sets of 50,000 scenarios each. For
these 20 sets, we now check how stable the resulting risk charge for
the risk factors is, in particular whether the shift (up- and down)
in each of the sets is within the gray boxes from the EIOPA MCRCS
study used above.

\begin{figure}[H]
\begin{centering}
\includegraphics[width=10cm]{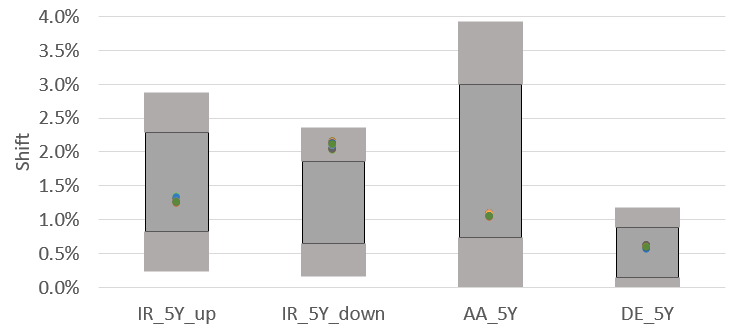}
\par\end{centering}
\caption{\label{fig:Comparison-stability}Comparison of the GAN risk charges
in 20 different runs versus MCRCS study for four risk factors}
\end{figure}

Figure \ref{fig:Comparison-stability} illustrates this for four different
risk factors, namely 5-years interest rate up- and down shift, 5-years
corporate credit spreads AA and 5-years German sovereign credit spreads.
For interest rate down shifts, we display the absolute values in the
graph. We here present the gray boxes from MCRCS study as above, but
instead of one blue dot for the GAN result, we show here 20 coloured
dots - each dot representing one of the 20 different runs. As the
dots overlap each other, not all dots can be seen in this graph.

For all the risk factors, even those that are not shown in the figure,
we analyzed that all shifts are within the gray boxes for every of
the 20 runs, except for the outliers in interest rates as commented
on in Section \ref{subsec:Comparison-on-risk-factor}.

For a quantitative stability check, we first calculate for each of
the 46 risk factors in each of the 20 scenario sets, the 0.5-percentile
and the 99.5-percentile (equalling the up- and down shift used in
MCRCS study). Then, for each risk factor and each percentile, we compute
the empirical first and third quartiles Q1 and Q3 over the 20 runs
and report the \emph{coefficient of quartile variation} by 
\[
CQV=\frac{Q_{3}-Q_{1}}{Q_{3}+Q_{1}},
\]

see \citet[eq. (2.30)]{kokoska2000crc}. Appendix \ref{sec:Appendix-2-Ticker}
shows the whole table of the results for all risk factors and both
percentiles. We can state that the coefficient of quartile variation
for both percentiles differs between 0.5\% and 10.0\%. This indicates
that the results are stable over different runs of the GAN.\\

Overall, the GAN-based model shows in every dimension in the study
comparable results to the certified internal models in Europe. Therefore,
the proof of concept of whether a GAN can serve as an ESG for market
risk modeling is successful. Please note that from a regulatory perspective,
it is desirable that all approved internal models for market risk
lead to comparable results. Therefore, this research indicates that
a GAN-based model can be seen as an appropriate alternative way of
market risk modeling.

\section{Conclusions and discussion of results}

In this research, we have shown how a generative adversarial network
(GAN) can serve as an economic scenario generator (ESG) for the calculation
of market risk in insurance companies. We used data from Bloomberg
to model financial instruments and to derive financial scenarios how
they can behave over a one year time horizon. We applied this to the
risk factors and benchmark portfolios of the EIOPA MCRCS study which
reflect typical market risk profiles of European insurance undertakings.
We have shown that the results of the GAN-based model are comparable
to the currently used ESGs which are usually based on Monte-Carlo
simulation using financial mathematical models. Hence our research
indicates that this approach could also serve as a regulatory approved
model as it performs well in the EIOPA benchmark study.

Compared to current approaches, a GAN-based ESG approach does not
require assumptions about the development of the risk factors, e.g.
on the drift of equities or on the negativity of interest rates. The
only assumption in a GAN-based model is that ``it is all in the data''.
This is similar to the assumption made in the calibration of a traditional
ESG where the empirical returns in the past serve as calibration targets.

The dependencies in a GAN-based model are automatically retrieved
from the empirical data. In modern risk models, dependencies are often
modeled using copulas. A certain copula has to be selected for each
dependency and must be fitted so that the desired dependencies are
reached. In practice, the modeling of dependencies is very difficult.

Calibration of the financial models to match the empirical data is
a task that has to be performed regularly by risk managers to keep
the models up to date. This is a cumbersome process and there is no
standard process for calibration, see \citet[Chapter 2.1]{DAV_Kalibrierung}.
This task is not needed for GAN-based models which makes them easier
to use. If new data are to be included, the GAN simply has to be fed
with the new data. Once the configuration and hyperparameter optimization
of the GAN has been set up, the training process is fairly straightforward.

One drawback of a GAN-based ESG is the fact that it relies purely
on events that have happened in the past in the financial markets
and cannot e.g. produce new dependencies that are not included in
the data the model is trained with. Classical financial models aim
to derive a theory based on developments in the past and can therefore
probably produce scenarios a GAN cannot come up with. Moreover, the
classical models are easier to interpret and explain to a Board or
a regulator whereas a GAN can be considered a ``black box''.

However, it is probable that a GAN-based ESG adapts faster to a regime-switch
in one of the model's risk factors: A classical ESG requires a new
financial model to be developed and implemented, whereas a GAN has
just to be trained with new data. To some extent, this behaviour can
be obtained in classical ESGs, too, if new data is weighted more heavily
compared to data longer ago.

In summary, a GAN-based internal market risk model is feasible and
can be seen as either an alternative to classical internal models
or as a benchmark.

\subsection*{Author Contributions}

Conceptualization, S.F.; methodology, S.F.; software, S.F.; investigation,
S.F. and G.J.; data analysis, S.F.; writing---original draft preparation,
S.F.; writing---review and editing, G.J.; visualization, S.F..; supervision,
G.J. All authors have read and agreed to the published version of
the manuscript.

\subsection*{Funding}

S. Flaig would like to thank Deutsche Rueckversicherung AG for the
funding of this research. Opinions, errors and omissions are solely
those of the authors and do not represent those of Deutsche Rueckversicherung
AG or its affiliates.

\subsection*{Data Availability Statement}

All the financial data used in this research is available via Bloomberg
with the tickers provided in the appendix.

\subsection*{Conflicts of interest}

The authors declare no conflict of interest.

\bibliographystyle{plainnat}
\phantomsection\addcontentsline{toc}{section}{\refname}\bibliography{8N__Gemeinsam_Marktrisiko_Flaig_DataScientist_Promotion_Promotionsdateien_Literatur_Bibtex}

\begin{thebibliography}{49}
\providecommand{\natexlab}[1]{#1}
\providecommand{\url}[1]{\texttt{#1}}
\expandafter\ifx\csname urlstyle\endcsname\relax
  \providecommand{\doi}[1]{doi: #1}\else
  \providecommand{\doi}{doi: \begingroup \urlstyle{rm}\Url}\fi

\bibitem[Aggarwal et~al.(2021)Aggarwal, Mittal, and
  Battineni]{aggarwal2021generative}
Alankrita Aggarwal, Mamta Mittal, and Gopi Battineni.
\newblock Generative adversarial network: An overview of theory and
  applications.
\newblock \emph{International Journal of Information Management Data Insights},
  page 100004, 2021.

\bibitem[Albrecht and Maurer(2016)]{albrecht2016investment}
Peter Albrecht and Raimond Maurer.
\newblock \emph{Investment-und Risikomanagement: Modelle, Methoden,
  Anwendungen}.
\newblock Sch{\"a}ffer-Poeschel, 2016.

\bibitem[Bennemann(2011)]{bennemann2011handbuch}
Christoph Bennemann.
\newblock \emph{{Handbuch Solvency II: von der Standardformel zum internen
  Modell, vom Governance-System zu den MaRisk VA}}.
\newblock Sch{\"a}ffer-Poeschel, 2011.

\bibitem[Borji(2019)]{borji2019pros}
Ali Borji.
\newblock Pros and cons of gan evaluation measures.
\newblock \emph{Computer Vision and Image Understanding}, 179:\penalty0 41--65,
  2019.

\bibitem[Chen et~al.(2018)Chen, Li, and Zhang]{chen2018bayesian}
Yize Chen, Pan Li, and Baosen Zhang.
\newblock Bayesian renewables scenario generation via deep generative networks.
\newblock In \emph{2018 52nd Annual Conference on Information Sciences and
  Systems (CISS)}, pages 1--6. IEEE, 2018.

\bibitem[Chollet(2018)]{chollet2018deep}
Francois Chollet.
\newblock \emph{Deep learning with Python}.
\newblock Manning, 2018.

\bibitem[Cote et~al.(2020)Cote, Hartman, Mercier, Meyers, Cummings, and
  Harmon]{cote2020synthesizing}
Marie-Pier Cote, Brian Hartman, Olivier Mercier, Joshua Meyers, Jared Cummings,
  and Elijah Harmon.
\newblock Synthesizing property \& casualty ratemaking datasets using
  generative adversarial networks.
\newblock \emph{arXiv preprint arXiv:2008.06110}, 2020.

\bibitem[Danthine(2017)]{danthine2017interest}
Jean-Pierre Danthine.
\newblock The interest rate unbound?
\newblock \emph{Comparative Economic Studies}, 59\penalty0 (2):\penalty0
  129--148, 2017.

\bibitem[{DAV (Deutsche Aktuarsvereinigung e.V.)}(2015)]{DAV_Kalibrierung}
{DAV (Deutsche Aktuarsvereinigung e.V.)}.
\newblock {Zwischenbericht zur Kalibrierung und Validierung spezieller ESG
  unter Solvency II. Ergebnisbericht des Ausschusses Investment der Deutschen
  Aktuarvereinigung e.V.}, 2015.
\newblock URL
  \url{https://aktuar.de/unsere-themen/fachgrundsaetze-oeffentlich/2015-11-09_DAV-Ergebnisbericht_Kalibrierung%20und%20Validierung%20spezieller%20ESG_Update.pdf}.

\bibitem[Denuit et~al.(2006)Denuit, Dhaene, Goovaerts, and
  Kaas]{denuit2006actuarial}
Michel Denuit, Jan Dhaene, Marc Goovaerts, and Rob Kaas.
\newblock \emph{Actuarial theory for dependent risks: measures, orders and
  models}.
\newblock John Wiley \& Sons, 2006.

\bibitem[Deutsch(2004)]{deutsch2004derivate}
Hans-Peter Deutsch.
\newblock \emph{{Derivate und interne Modelle: modernes Risikomanagement}}.
\newblock Sch{\"a}ffer-Poeschel, 3rd edition, 2004.

\bibitem[Eckerli and Osterrieder(2021)]{eckerli2021generative}
Florian Eckerli and Joerg Osterrieder.
\newblock Generative adversarial networks in finance: an overview.
\newblock \emph{arXiv preprint arXiv:2106.06364}, 2021.

\bibitem[{EIOPA}(2014)]{eiopa2014underlying}
{EIOPA}.
\newblock The underlying assumptions in the standard formula for the solvency
  capital requirement calculation, 2014.
\newblock URL
  \url{https://www.bafin.de/SharedDocs/Downloads/EN/Leitfaden/VA/dl_lf_solvency_annahmen_standardformel_scr_en.pdf}.

\bibitem[EIOPA(2019)]{EIOPA_Zinskurve_TechnicalDocumentation}
EIOPA.
\newblock Technical documentation of the methodology to derive eiopas risk-free
  interest rate term structures, 2019.
\newblock URL
  \url{https://www.eiopa.europa.eu/sites/default/files/risk_free_interest_rate/12092019{\-}technical_documentation.pdf}.

\bibitem[{EIOPA}(2021{\natexlab{a}})]{EIOPA_MCRCS_Ergebnis}
{EIOPA}.
\newblock {YE2019 Comparative Study on Market \& Credit Risk Modelling},
  2021{\natexlab{a}}.
\newblock URL
  \url{https://www.eiopa.europa.eu/sites/default/files/publications/reports/2021-study-on-modelling-of-market-and-credit-risk-_mcrcs.pdf}.

\bibitem[{EIOPA}(2021{\natexlab{b}})]{eiopa2021insuranceoverview}
{EIOPA}.
\newblock European insurance overview 2021, 2021{\natexlab{b}}.
\newblock URL
  \url{https://www.eiopa.europa.eu/sites/default/files/publications/reports/eiopa-21-591-european-insurance-overview-report.pdf}.

\bibitem[{EIOPA MCRCS Project Group}(2020{\natexlab{a}})]{EIOPA_MCRCS_Excel}
{EIOPA MCRCS Project Group}.
\newblock {Specification of financial instruments and benchmark portfolios of
  the Year-end 2019 edition of the Market and credit risk modelling comparative
  study}, 2020{\natexlab{a}}.
\newblock URL
  \url{https://www.eiopa.europa.eu/sites/default/files/toolsanddata/mcrcs_2019_instruments_and_bmp.xlsx}.

\bibitem[{EIOPA MCRCS Project
  Group}(2020{\natexlab{b}})]{EIOPA_MCRCS_Instructions}
{EIOPA MCRCS Project Group}.
\newblock {Market \& credit risk modelling comparative study (MCRCS), year-end
  2019 edition: Instructions to participating undertakings for filling out the
  data request}, 2020{\natexlab{b}}.
\newblock URL
  \url{https://www.eiopa.europa.eu/sites/default/files/toolsanddata/mcrcs_year-end_2019_instructions_covidpostponed.pdf}.

\bibitem[{European Commission}(2015)]{delegated2015commission}
{European Commission}.
\newblock {Commission delegated regulation (EU) 2015/35 of 10 October 2014
  supplementing directive 2009/138/EC of the European parliament and of the
  council on the taking-up and pursuit of the business of insurance and
  reinsurance (Solvency II)}.
\newblock \emph{Official Journal of European Union}, 2015.

\bibitem[Franco-Pedroso et~al.(2019)Franco-Pedroso, Gonzalez-Rodriguez, Cubero,
  Planas, Cobo, and Pablos]{franco2019generating}
Javier Franco-Pedroso, Joaquin Gonzalez-Rodriguez, Jorge Cubero, Maria Planas,
  Rafael Cobo, and Fernando Pablos.
\newblock Generating virtual scenarios of multivariate financial data for
  quantitative trading applications.
\newblock \emph{The Journal of Financial Data Science}, 1\penalty0
  (2):\penalty0 55--77, 2019.

\bibitem[Fu et~al.(2019)Fu, Chen, Zeng, Zhuang, and Sudjianto]{fu2019time}
Rao Fu, Jie Chen, Shutian Zeng, Yiping Zhuang, and Agus Sudjianto.
\newblock Time series simulation by conditional generative adversarial net.
\newblock \emph{arXiv preprint arXiv:1904.11419}, 2019.

\bibitem[Goodfellow(2016)]{goodfellow2016nips}
Ian Goodfellow.
\newblock {Nips 2016 tutorial: Generative adversarial networks}.
\newblock \emph{arXiv preprint arXiv:1701.00160}, 2016.

\bibitem[Goodfellow et~al.(2014)Goodfellow, Pouget-Abadie, and
  Mirza]{goodfellow2014pouget}
Ian Goodfellow, J.~Pouget-Abadie, and M.~et~al Mirza.
\newblock Generative adversarial nets.
\newblock \emph{Advances in neural information processing systems}, 2014.

\bibitem[Grasselli and Lipton(2019)]{grasselli2019normality}
Matheus~R Grasselli and Alexander Lipton.
\newblock On the normality of negative interest rates.
\newblock \emph{Review of Keynesian Economics}, 7\penalty0 (2):\penalty0
  201--219, 2019.

\bibitem[Gr{\"u}ndl et~al.(2019)Gr{\"u}ndl, Kraft, Post, Schulze, Pelzer, and
  Schl{\"u}tter]{grundl2019solvency}
Helmut Gr{\"u}ndl, Mirko Kraft, Thomas Post, Roman~N Schulze, Sabine Pelzer,
  and Sebastian Schl{\"u}tter.
\newblock \emph{Solvency II-Eine Einf{\"u}hrung: Grundlagen der neuen
  Versicherungsaufsicht}.
\newblock VVW GmbH, 2nd edition, 2019.

\bibitem[Hallin et~al.(2021)Hallin, Mordant, and
  Segers]{hallin2021multivariate}
Marc Hallin, Gilles Mordant, and Johan Segers.
\newblock Multivariate goodness-of-fit tests based on wasserstein distance.
\newblock \emph{Electronic Journal of Statistics}, 15\penalty0 (1):\penalty0
  1328--1371, 2021.

\bibitem[Henry-Labordere(2019)]{henry2019generative}
Pierre Henry-Labordere.
\newblock Generative models for financial data.
\newblock \emph{Available at SSRN 3408007}, 2019.

\bibitem[Ho and Wookey(2019)]{ho2019real}
Yaoshiang Ho and Samuel Wookey.
\newblock The real-world-weight cross-entropy loss function: Modeling the costs
  of mislabeling.
\newblock \emph{IEEE Access}, 8:\penalty0 4806--4813, 2019.

\bibitem[Kokoska and Zwillinger(2000)]{kokoska2000crc}
Stephen Kokoska and Daniel Zwillinger.
\newblock \emph{CRC standard probability and statistics tables and formulae}.
\newblock Crc Press, 2000.

\bibitem[Kuo(2019)]{kuo2019generative}
Kevin Kuo.
\newblock Generative synthesis of insurance datasets.
\newblock \emph{arXiv preprint arXiv:1912.02423}, 2019.

\bibitem[Lager{\aa}s and Lindholm(2016)]{lageraas2016issues}
Andreas Lager{\aa}s and Mathias Lindholm.
\newblock Issues with the smith--wilson method.
\newblock \emph{Insurance: Mathematics and Economics}, 71:\penalty0 93--102,
  2016.

\bibitem[Lezmi et~al.(2020)Lezmi, Roche, Roncalli, and Xu]{lezmi2020improving}
Edmond Lezmi, Jules Roche, Thierry Roncalli, and Jiali Xu.
\newblock Improving the robustness of trading strategy backtesting with
  boltzmann machines and generative adversarial networks.
\newblock \emph{Available at SSRN 3645473}, 2020.

\bibitem[Li et~al.(2020)Li, Tao, and Li]{li2020regularization}
Ziqiang Li, Rentuo Tao, and Bin Li.
\newblock Regularization and normalization for generative adversarial networks:
  A review.
\newblock \emph{arXiv preprint arXiv:2008.08930}, 2020.

\bibitem[Marti(2020)]{marti2020corrgan}
Gautier Marti.
\newblock Corrgan: Sampling realistic financial correlation matrices using
  generative adversarial networks.
\newblock In \emph{ICASSP 2020-2020 IEEE International Conference on Acoustics,
  Speech and Signal Processing (ICASSP)}, pages 8459--8463. IEEE, 2020.

\bibitem[Mazumdar et~al.(2020)Mazumdar, Ratliff, and
  Sastry]{mazumdar2020gradient}
Eric Mazumdar, Lillian~J Ratliff, and S~Shankar Sastry.
\newblock On gradient-based learning in continuous games.
\newblock \emph{SIAM Journal on Mathematics of Data Science}, 2\penalty0
  (1):\penalty0 103--131, 2020.

\bibitem[Motwani and Parmar(2020)]{motwani2020novel}
Tanya Motwani and Manojkumar Parmar.
\newblock {A novel framework for selection of GANs for an application}.
\newblock \emph{arXiv preprint arXiv:2002.08641}, 2020.

\bibitem[Ngwenduna and Mbuvha(2021)]{ngwenduna2021alleviating}
Kwanda~Sydwell Ngwenduna and Rendani Mbuvha.
\newblock Alleviating class imbalance in actuarial applications using
  generative adversarial networks.
\newblock \emph{Risks}, 9\penalty0 (3):\penalty0 49, 2021.

\bibitem[Ni et~al.(2020)Ni, Szpruch, Wiese, Liao, and Xiao]{ni2020conditional}
Hao Ni, Lukasz Szpruch, Magnus Wiese, Shujian Liao, and Baoren Xiao.
\newblock Conditional sig-wasserstein gans for time series generation.
\newblock \emph{arXiv preprint arXiv:2006.05421}, 2020.

\bibitem[Pedersen et~al.(2016)Pedersen, Campbell, Christiansen, Cox, Finn,
  Griffin, Hooker, Lightwood, Sonlin, and Suchar]{pedersen2016economic}
Hal Pedersen, Mary~Pat Campbell, Stephan~L Christiansen, Samuel~H Cox, Daniel
  Finn, Ken Griffin, Nigel Hooker, Matthew Lightwood, Stephen~M Sonlin, and
  Chris Suchar.
\newblock Economic scenario generators: A practical guide.
\newblock \emph{The Society of Actuaries (July 2016)}, 2016.

\bibitem[Pfeifer and Ragulina(2018)]{pfeifer2018generating}
Dietmar Pfeifer and Olena Ragulina.
\newblock {Generating VaR scenarios under Solvency II with product beta
  distributions}.
\newblock \emph{Risks}, 6\penalty0 (4):\penalty0 122, 2018.

\bibitem[Ratings(2018)]{SP_RatingTransistions}
S{\&}P Ratings.
\newblock Annual global corporate default and rating transition study, 2018.
\newblock URL
  \url{https://www.spratings.com/documents/20184/774196/2018AnnualGlobalCorporateDefaultAndRatingTransitionStudy.pdf}.

\bibitem[Salimans et~al.(2016)Salimans, Goodfellow, Zaremba, Cheung, Radford,
  and Chen]{salimans2016improved}
Tim Salimans, Ian Goodfellow, Wojciech Zaremba, Vicki Cheung, Alec Radford, and
  Xi~Chen.
\newblock Improved techniques for training gans.
\newblock \emph{arXiv preprint arXiv:1606.03498}, 2016.

\bibitem[Smith and Wilson(2001)]{smith2001fitting}
Andrew Smith and Tim Wilson.
\newblock Fitting yield curves with long term constraints.
\newblock Technical report, Technical report, 2001.

\bibitem[Theis et~al.(2015)Theis, Oord, and Bethge]{theis2015note}
Lucas Theis, A{\"a}ron van~den Oord, and Matthias Bethge.
\newblock A note on the evaluation of generative models.
\newblock \emph{arXiv preprint arXiv:1511.01844}, 2015.

\bibitem[Viehmann(2019)]{viehmann2019variants}
Thomas Viehmann.
\newblock Variants of the smith-wilson method with a view towards applications.
\newblock \emph{arXiv preprint arXiv:1906.06363}, 2019.

\bibitem[Wiese et~al.(2019)Wiese, Bai, Wood, and Buehler]{wiese2019deep}
Magnus Wiese, Lianjun Bai, Ben Wood, and Hans Buehler.
\newblock Deep hedging: learning to simulate equity option markets.
\newblock \emph{arXiv preprint arXiv:1911.01700}, 2019.

\bibitem[Wiese et~al.(2020)Wiese, Knobloch, Korn, and
  Kretschmer]{wiese2020quant}
Magnus Wiese, Robert Knobloch, Ralf Korn, and Peter Kretschmer.
\newblock Quant gans: Deep generation of financial time series.
\newblock \emph{Quantitative Finance}, 20\penalty0 (9):\penalty0 1419--1440,
  2020.

\bibitem[Yoon et~al.(2019)Yoon, Jarrett, and Van~der Schaar]{yoon2019time}
Jinsung Yoon, Daniel Jarrett, and Mihaela Van~der Schaar.
\newblock Time-series generative adversarial networks.
\newblock \emph{Advances in neural information processing systems}, 32, 2019.

\bibitem[Yu(2002)]{yu2002resampling}
Chong~Ho Yu.
\newblock Resampling methods: concepts, applications, and justification.
\newblock \emph{Practical Assessment, Research, and Evaluation}, 8\penalty0
  (1):\penalty0 19, 2002.

\end{thebibliography}

\appendix

\subsection*{Appendix}

\section{\label{sec:Appendix-2-Ticker}Table of risk factors and data sources
used for the MCRCS study}

Here is the ticker list from Bloomberg for the data used for the 46
risk factors used in our GAN training. Additionally, the coefficient
of quartile variation $CQV$ for up- and down-shifts for all risk
factors over the 20 different GAN runs as described in Section \ref{subsec:Stability-of-GAN}
are shown.

\begin{table}[H]
\centering{}
\end{table}

\begin{center}
\begin{longtable}[c]{|>{\centering}p{2.5cm}|>{\centering}p{3cm}|>{\centering}p{1.5cm}|>{\centering}p{4cm}|>{\centering}p{1.25cm}|>{\centering}p{1.25cm}|}
\hline 
{\small{}Asset Class} &
{\small{}Subtype} &
{\small{}Maturity} &
{\small{}Bloomberg ticker} &
{\small{}$CQV$, 0.5perc.} &
{\small{}$CQV$, 99.5perc.}\tabularnewline
\hline 
\endhead
\hline 
{\small{}Government bond} &
{\small{}Austria} &
{\small{}5 years} &
{\small{}GTATS5Y Govt} &
{\small{}9.2\%} &
{\small{}4.8\%}\tabularnewline
\hline 
 &
{\small{}Austria} &
{\small{}10 years} &
{\small{}GTATS10Y Govt} &
{\small{}4.3\%} &
{\small{}3.2\%}\tabularnewline
\hline 
 &
{\small{}Belgium} &
{\small{}5 years} &
{\small{}GTBEF5Y Govt} &
{\small{}4.1\%} &
{\small{}4.0\%}\tabularnewline
\hline 
 &
{\small{}Belgium} &
{\small{}10 years} &
{\small{}GTBEF10Y Govt} &
{\small{}5.5\%} &
{\small{}5.7\%}\tabularnewline
\hline 
 &
{\small{}Germany} &
{\small{}5 years} &
{\small{}GTDEM5Y Govt} &
{\small{}1.4\%} &
{\small{}4.8\%}\tabularnewline
\hline 
 &
{\small{}Germany} &
{\small{}10 years} &
{\small{}GTDEM10Y Govt} &
{\small{}3.3\%} &
{\small{}4.5\%}\tabularnewline
\hline 
 &
{\small{}Spain} &
{\small{}5 years} &
{\small{}GTESP5Y Govt} &
{\small{}10.0\%} &
{\small{}1.9\%}\tabularnewline
\hline 
 &
{\small{}Spain} &
{\small{}10 years} &
{\small{}GTESP10Y Govt} &
{\small{}2.5\%} &
{\small{}4.0\%}\tabularnewline
\hline 
 &
{\small{}France} &
{\small{}5 years} &
{\small{}GTFRF5Y Govt} &
{\small{}2.3\%} &
{\small{}9.5\%}\tabularnewline
\hline 
 &
{\small{}France} &
{\small{}10 years} &
{\small{}GTFRF10Y Govt} &
{\small{}3.6\%} &
{\small{}4.8\%}\tabularnewline
\hline 
 &
{\small{}Ireland} &
{\small{}5 years} &
{\small{}GIGB5Y Index} &
{\small{}2.7\%} &
{\small{}4.0\%}\tabularnewline
\hline 
 &
{\small{}Ireland} &
{\small{}10 years} &
{\small{}GIGB10Y Index} &
{\small{}2.2\%} &
{\small{}7.8\%}\tabularnewline
\hline 
 &
{\small{}Italy} &
{\small{}5 years} &
{\small{}GTITL5Y Govt} &
{\small{}4.0\%} &
{\small{}8.0\%}\tabularnewline
\hline 
 &
{\small{}Italy} &
{\small{}10 years} &
{\small{}GTITL10Y Govt} &
{\small{}2.0\%} &
{\small{}2.7\%}\tabularnewline
\hline 
 &
{\small{}Netherlands} &
{\small{}5 years} &
{\small{}GTNLG5Y Govt} &
{\small{}4.8\%} &
{\small{}1.4\%}\tabularnewline
\hline 
 &
{\small{}Netherlands} &
{\small{}10 years} &
{\small{}GTNLG10Y Govt} &
{\small{}3.1\%} &
{\small{}6.5\%}\tabularnewline
\hline 
 &
{\small{}Portugal} &
{\small{}5 years} &
{\small{}GSPT5YR Index} &
{\small{}3.1\%} &
{\small{}3.3\%}\tabularnewline
\hline 
 &
{\small{}UK} &
{\small{}5 years} &
{\small{}C1105Y Index} &
{\small{}5.6\%} &
{\small{}4.3\%}\tabularnewline
\hline 
 &
{\small{}US} &
{\small{}5 years} &
{\small{}H15T5Y Index} &
{\small{}3.0\%} &
{\small{}5.3\%}\tabularnewline
\hline 
{\small{}Covered bond} &
{\small{}AA-rated issuer} &
{\small{}5 years} &
{\small{}C9235Y Index} &
{\small{}2.6\%} &
{\small{}4.3\%}\tabularnewline
\hline 
 &
{\small{}AA-rated issuer} &
{\small{}10 years} &
{\small{}C92310Y Index} &
{\small{}3.2\%} &
{\small{}4.3\%}\tabularnewline
\hline 
{\small{}Corporate bond} &
{\small{}bond, rating AA} &
{\small{}5 years} &
{\small{}C6675Y Index} &
{\small{}1.9\%} &
{\small{}4.4\%}\tabularnewline
\hline 
 &
{\small{}bond, rating AA} &
{\small{}10 years} &
{\small{}C66710Y Index} &
{\small{}2.8\%} &
{\small{}1.6\%}\tabularnewline
\hline 
 &
{\small{}bond, rating A} &
{\small{}5 years} &
{\small{}C6705Y Index} &
{\small{}1.6\%} &
{\small{}3.2\%}\tabularnewline
\hline 
 &
{\small{}bond, rating A} &
{\small{}10 years} &
{\small{}C67010Y Index} &
{\small{}3.5\%} &
{\small{}2.5\%}\tabularnewline
\hline 
 &
{\small{}bond, rating BBB} &
{\small{}5 years} &
{\small{}C6735Y Index} &
{\small{}5.5\%} &
{\small{}5.0\%}\tabularnewline
\hline 
 &
{\small{}bond, rating BBB} &
{\small{}10 years} &
{\small{}C67310Y Index} &
{\small{}4.9\%} &
{\small{}3.4\%}\tabularnewline
\hline 
 &
{\small{}high yield bonds} &
{\small{}5 years} &
{\small{}ML HP00 Swap Spread} &
{\small{}1.4\%} &
{\small{}4.4\%}\tabularnewline
\hline 
{\small{}Interest rates, risk-free} &
{\small{}EUR} &
{\small{}1 year} &
{\small{}S0045Z 1Y BLC2 Curncy} &
{\small{}1.9\%} &
{\small{}0.5\%}\tabularnewline
\hline 
 &
{\small{}EUR} &
{\small{}3 years} &
{\small{}S0045Z 3Y BLC2 Curncy} &
{\small{}2.6\%} &
{\small{}2.0\%}\tabularnewline
\hline 
 &
{\small{}EUR} &
{\small{}5 years} &
{\small{}S0045Z 5Y BLC2 Curncy} &
{\small{}2.1\%} &
{\small{}1.0\%}\tabularnewline
\hline 
 &
{\small{}EUR} &
{\small{}7 years} &
{\small{}S0045Z 7Y BLC2 Curncy} &
{\small{}3.5\%} &
{\small{}1.0\%}\tabularnewline
\hline 
 &
{\small{}EUR} &
{\small{}10 years} &
{\small{}S0045Z 10Y BLC2 Curncy} &
{\small{}4.8\%} &
{\small{}2.6\%}\tabularnewline
\hline 
 &
{\small{}EUR} &
{\small{}15 years} &
{\small{}S0045Z 15Y BLC2 Curncy} &
{\small{}4.4\%} &
{\small{}3.2\%}\tabularnewline
\hline 
 &
{\small{}EUR} &
{\small{}20 years} &
{\small{}S0045Z 20Y BLC2 Curncy} &
{\small{}3.0\%} &
{\small{}1.6\%}\tabularnewline
\hline 
 &
{\small{}EUR} &
{\small{}25 years} &
{\small{}S0045Z 25Y BLC2 Curncy} &
{\small{}2.2\%} &
{\small{}3.3\%}\tabularnewline
\hline 
 &
{\small{}EUR} &
{\small{}30 years} &
{\small{}S0045Z 30Y BLC2 Curncy} &
{\small{}4.6\%} &
{\small{}2.4\%}\tabularnewline
\hline 
 &
{\small{}EUR} &
{\small{}40 years} &
{\small{}S0045Z 40Y BLC2 Curncy} &
{\small{}4.4\%} &
{\small{}2.5\%}\tabularnewline
\hline 
 &
{\small{}EUR} &
{\small{}50 years} &
{\small{}S0045Z 50Y BLC2 Curncy} &
{\small{}1.7\%} &
{\small{}3.2\%}\tabularnewline
\hline 
 &
{\small{}USD} &
{\small{}5 years} &
{\small{}USSW5 Index} &
{\small{}1.4\%} &
{\small{}2.2\%}\tabularnewline
\hline 
 &
{\small{}GBP} &
{\small{}5 years} &
{\small{}BPSW5 Index} &
{\small{}4.6\%} &
{\small{}3.3\%}\tabularnewline
\hline 
{\small{}Equity} &
{\small{}EuroStoxx50} &
{\small{}-} &
{\small{}SX5T Index} &
{\small{}6.1\%} &
{\small{}5.6\%}\tabularnewline
\hline 
 &
{\small{}MSCI Europe} &
{\small{}-} &
{\small{}MSDEE15N Index} &
{\small{}4.2\%} &
{\small{}3.0\%}\tabularnewline
\hline 
 &
{\small{}FTSE100} &
{\small{}-} &
{\small{}TUKXG Index} &
{\small{}4.1\%} &
{\small{}3.2\%}\tabularnewline
\hline 
 &
{\small{}S\&P500} &
{\small{}-} &
{\small{}SPTR500N Index} &
{\small{}4.7\%} &
{\small{}6.8\%}\tabularnewline
\hline 
{\small{}Real-estate} &
{\small{}Europe, commercial} &
{\small{}-} &
{\small{}EXUK Index} &
{\small{}2.8\%} &
{\small{}4.5\%}\tabularnewline
\hline 
\end{longtable}
\par\end{center}

\begin{center}
\begin{table}[H]
\caption{\label{tab:Mapping-of-risk-factors-CQV}Risk factors, Bloomberg data
source and CQV used in the GAN-based internal model}
\end{table}
\par\end{center}

\section{\label{sec:Appendix-2-Instruments}Table of instruments used for
the MCRCS study}

In this table, all instruments used for comparison with MCRCS study
in the paper are displayed.

\begin{longtable}[c]{|c|c|>{\centering}p{7cm}|}
\hline 
{\small{}Instrument} &
{\small{}Maturity} &
{\small{}Risk factors used for valuation}\tabularnewline
\hline 
\endhead
\hline 
{\small{}EUR risk-free interest rate} &
{\small{}1y} &
{\small{}EUR swap rate, 1y}\tabularnewline
\hline 
{\small{}EUR risk-free interest rate} &
{\small{}3y} &
{\small{}EUR swap rate, 3y}\tabularnewline
\hline 
{\small{}EUR risk-free interest rate} &
{\small{}5y} &
{\small{}EUR swap rate, 5y}\tabularnewline
\hline 
{\small{}EUR risk-free interest rate} &
{\small{}7y} &
{\small{}EUR swap rate, 7y}\tabularnewline
\hline 
{\small{}EUR risk-free interest rate} &
{\small{}10y} &
{\small{}EUR swap rate, 10y}\tabularnewline
\hline 
{\small{}EUR risk-free interest rate} &
{\small{}15y} &
{\small{}EUR swap rate, 15y}\tabularnewline
\hline 
{\small{}EUR risk-free interest rate} &
{\small{}20y} &
{\small{}EUR swap rate, 20y}\tabularnewline
\hline 
{\small{}EUR risk-free interest rate} &
{\small{}25y} &
{\small{}EUR swap rate, 25y}\tabularnewline
\hline 
{\small{}EUR risk-free interest rate} &
{\small{}30y} &
{\small{}EUR swap rate, 30y}\tabularnewline
\hline 
{\small{}EUR risk-free interest rate} &
{\small{}40y} &
{\small{}EUR swap rate, 40y}\tabularnewline
\hline 
{\small{}EUR risk-free interest rate} &
{\small{}50y} &
{\small{}EUR swap rate, 50y}\tabularnewline
\hline 
{\small{}EUR risk-free interest rate} &
{\small{}60y} &
{\small{}EUR swap rate, 60y}\tabularnewline
\hline 
{\small{}Austrian Sovereign bond} &
{\small{}5y} &
{\small{}EUR interest rate, 5y \& AT\_Spread, 5y}\tabularnewline
\hline 
{\small{}Austrian Sovereign bond} &
{\small{}10y} &
{\small{}EUR interest rate, 10y \& AT\_Spread, 10y}\tabularnewline
\hline 
{\small{}Austrian Sovereign bond} &
{\small{}20y} &
{\small{}EUR interest rate, 20y \& AT\_Spread, 10y}\tabularnewline
\hline 
{\small{}Belgium Sovereign bond} &
{\small{}5y} &
{\small{}EUR interest rate, 5y \& BE\_Spread, 5y}\tabularnewline
\hline 
{\small{}Belgium Sovereign bond} &
{\small{}10y} &
{\small{}EUR interest rate, 10y \& BE\_Spread, 10y}\tabularnewline
\hline 
{\small{}Belgium Sovereign bond} &
{\small{}20y} &
{\small{}EUR interest rate, 20y \& BE\_Spread, 10y}\tabularnewline
\hline 
{\small{}German Sovereign bond} &
{\small{}5y} &
{\small{}EUR interest rate, 5y \& DE\_Spread, 5y}\tabularnewline
\hline 
{\small{}German Sovereign bond} &
{\small{}10y} &
{\small{}EUR interest rate, 10y \& DE\_Spread, 10y}\tabularnewline
\hline 
{\small{}German Sovereign bond} &
{\small{}20y} &
{\small{}EUR interest rate, 20y \& DE\_Spread, 10y}\tabularnewline
\hline 
{\small{}Spain Sovereign bond} &
{\small{}5y} &
{\small{}EUR interest rate, 5y \& ES\_Spread, 5y}\tabularnewline
\hline 
{\small{}Spain Sovereign bond} &
{\small{}10y} &
{\small{}EUR interest rate, 10y \& ES\_Spread, 10y}\tabularnewline
\hline 
{\small{}Spain Sovereign bond} &
{\small{}20y} &
{\small{}EUR interest rate, 20y \& ES\_Spread, 10y}\tabularnewline
\hline 
{\small{}France Sovereign bond} &
{\small{}5y} &
{\small{}EUR interest rate, 5y \& FR\_Spread, 5y}\tabularnewline
\hline 
{\small{}France Sovereign bond} &
{\small{}10y} &
{\small{}EUR interest rate, 10y \& FR\_Spread, 10y}\tabularnewline
\hline 
{\small{}France Sovereign bond} &
{\small{}20y} &
{\small{}EUR interest rate, 20y \& FR\_Spread, 10y}\tabularnewline
\hline 
{\small{}Ireland Sovereign bond} &
{\small{}5y} &
{\small{}EUR interest rate, 5y \& IE\_Spread, 5y}\tabularnewline
\hline 
{\small{}Ireland Sovereign bond} &
{\small{}10y} &
{\small{}EUR interest rate, 10y \& IE\_Spread, 10y}\tabularnewline
\hline 
{\small{}Ireland Sovereign bond} &
{\small{}20y} &
{\small{}EUR interest rate, 20y \& IE\_Spread, 10y}\tabularnewline
\hline 
{\small{}Italia Sovereign bond} &
{\small{}5y} &
{\small{}EUR interest rate, 5y \& IT\_Spread, 5y}\tabularnewline
\hline 
{\small{}Italia Sovereign bond} &
{\small{}10y} &
{\small{}EUR interest rate, 10y \& IT\_Spread, 10y}\tabularnewline
\hline 
{\small{}Italia Sovereign bond} &
{\small{}20y} &
{\small{}EUR interest rate, 20y \& IT\_Spread, 10y}\tabularnewline
\hline 
{\small{}Netherlands Sovereign bond} &
{\small{}5y} &
{\small{}EUR interest rate, 5y \& NE\_Spread, 5y}\tabularnewline
\hline 
{\small{}Netherlands Sovereign bond} &
{\small{}10y} &
{\small{}EUR interest rate, 10y \& NE\_Spread, 10y}\tabularnewline
\hline 
{\small{}Netherlands Sovereign bond} &
{\small{}20y} &
{\small{}EUR interest rate, 20y \& NE\_Spread, 10y}\tabularnewline
\hline 
{\small{}Portugal Sovereign bond} &
{\small{}5y} &
{\small{}EUR interest rate, 5y \& PT\_Spread, 5y}\tabularnewline
\hline 
{\small{}UK Sovereign bond} &
{\small{}5y} &
{\small{}GBP interest rate, 5y \& UK\_Spread, 5y}\tabularnewline
\hline 
{\small{}US Sovereign bond} &
{\small{}5y} &
{\small{}USD interest rate, 5y \& US\_Spread, 5y}\tabularnewline
\hline 
{\small{}Bond issued by ESM} &
{\small{}10y} &
{\small{}EUR interest rate, 10y \& DE\_Spread, 10y}\tabularnewline
\hline 
{\small{}Covered bond rated AAA} &
{\small{}5y} &
{\small{}EUR interest rate, 5y \& COV\_Spread, 5y}\tabularnewline
\hline 
{\small{}Covered bond rated AAA} &
{\small{}10y} &
{\small{}EUR interest rate, 10y \& COV\_Spread, 10y}\tabularnewline
\hline 
{\small{}Financial bond, rated AAA} &
{\small{}5y} &
{\small{}EUR interest rate, 5y \& COV\_Spread, 5y}\tabularnewline
\hline 
{\small{}Financial bond, rated AAA} &
{\small{}10y} &
{\small{}EUR interest rate, 10y \& COV\_Spread, 10y}\tabularnewline
\hline 
{\small{}Financial bond, rated AA} &
{\small{}5y} &
{\small{}EUR interest rate, 5y \& AA\_Spread, 5y}\tabularnewline
\hline 
{\small{}Financial bond, rated AA} &
{\small{}10y} &
{\small{}EUR interest rate, 10y \& AA\_Spread, 10y}\tabularnewline
\hline 
{\small{}Financial bond, rated A} &
{\small{}5y} &
{\small{}EUR interest rate, 5y \& A\_Spread, 5y}\tabularnewline
\hline 
{\small{}Financial bond, rated A} &
{\small{}10y} &
{\small{}EUR interest rate, 10y \& A\_Spread, 10y}\tabularnewline
\hline 
{\small{}Financial bond, rated BBB} &
{\small{}5y} &
{\small{}EUR interest rate, 5y \& BBB\_Spread, 5y}\tabularnewline
\hline 
{\small{}Financial bond, rated BBB} &
{\small{}10y} &
{\small{}EUR interest rate, 10y \& BBB\_Spread, 10y}\tabularnewline
\hline 
{\small{}Financial bond, rated BB} &
{\small{}5y} &
{\small{}EUR interest rate, 5y \& HY\_Spread, 5y}\tabularnewline
\hline 
{\small{}Financial bond, rated BB} &
{\small{}10y} &
{\small{}EUR interest rate, 10y \& HY\_Spread, 10y}\tabularnewline
\hline 
{\small{}Non-Financial bond, rated AAA} &
{\small{}5y} &
{\small{}EUR interest rate, 5y \& COV\_Spread, 5y}\tabularnewline
\hline 
{\small{}Non-Financial bond, rated AAA} &
{\small{}10y} &
{\small{}EUR interest rate, 10y \& COV\_Spread, 10y}\tabularnewline
\hline 
{\small{}Non-Financial bond, rated AA} &
{\small{}5y} &
{\small{}EUR interest rate, 5y \& AA\_Spread, 5y}\tabularnewline
\hline 
{\small{}Non-Financial bond, rated AA} &
{\small{}10y} &
{\small{}EUR interest rate, 10y \& AA\_Spread, 10y}\tabularnewline
\hline 
{\small{}Non-Financial bond, rated A} &
{\small{}5y} &
{\small{}EUR interest rate, 5y \& A\_Spread, 5y}\tabularnewline
\hline 
{\small{}Non-Financial bond, rated A} &
{\small{}10y} &
{\small{}EUR interest rate, 10y \& A\_Spread, 10y}\tabularnewline
\hline 
{\small{}Non-Financial bond, rated BBB} &
{\small{}5y} &
{\small{}EUR interest rate, 5y \& BBB\_Spread, 5y}\tabularnewline
\hline 
{\small{}Non-Financial bond, rated BBB} &
{\small{}10y} &
{\small{}EUR interest rate, 10y \& BBB\_Spread, 10y}\tabularnewline
\hline 
{\small{}Non-Financial bond, rated BB} &
{\small{}5y} &
{\small{}EUR interest rate, 5y \& HY\_Spread, 5y}\tabularnewline
\hline 
{\small{}Non-Financial bond, rated BB} &
{\small{}10y} &
{\small{}EUR interest rate, 10y \& HY\_Spread, 10y}\tabularnewline
\hline 
{\small{}Equity Index, Eurostoxx 50} &
- &
{\small{}Equity Index, Eurostoxx 50}\tabularnewline
\hline 
{\small{}Equity Index, MSCI Europe} &
- &
{\small{}Equity Index, MSCI Europe}\tabularnewline
\hline 
{\small{}Equity Index, FTSE100} &
- &
{\small{}Equity Index, FTSE100}\tabularnewline
\hline 
{\small{}Equity Index, S\&P500} &
- &
{\small{}Equity Index, S\&P500}\tabularnewline
\hline 
{\small{}Residential real estate in Netherlands} &
- &
{\small{}Diversified European REIT index}\tabularnewline
\hline 
{\small{}Commercial real estate in France} &
- &
{\small{}Diversified European REIT index}\tabularnewline
\hline 
{\small{}Commercial real estate in Germany} &
- &
{\small{}Diversified European REIT index}\tabularnewline
\hline 
{\small{}Commercial real estate in UK} &
- &
{\small{}Diversified European REIT index}\tabularnewline
\hline 
{\small{}Commercial real estate in Italy} &
- &
{\small{}Diversified European REIT index}\tabularnewline
\hline 
\end{longtable}
\begin{center}
\begin{table}[H]
\caption{\label{tab:Mapping-of-instruments}Mapping of instruments to risk
factors used in the GAN-based internal model}
\end{table}
\par\end{center}

Please find here the explanations to the approximations and simplifications
used due to data availability reasons:
\begin{itemize}
\item As most participants in the study, we do not distinguish between different
types of corporate bond spreads, i.e. financial and non-financial
corporates are modeled with the same data. As written in \citet[p. 24]{EIOPA_MCRCS_Ergebnis},
this is a simplification used by two thirds of the participants.
\item As for the required supranational paper issued by ESM (European Stability
Mechanism), there is no long time series to be found, we use the approximation
of the German spreads instead.
\item There is no reliable daily data source for AAA and high yield bonds
in Bloomberg. For AAA rated bonds, as most participants in the study,
we use the covered bond spreads which are also rated AAA instead see
\citet[p. 24]{EIOPA_MCRCS_Ergebnis}. The most frequent data for high
yield bonds that we found can be derived from the Meryll Lynch spread
index which is a weekly index.
\item For real estate, there is no direct transaction based data available
on high frequency. The most frequent direct real estate data is available
on a monthly basis. We will therefore use an index representing Real
Estate Investment Trusts (REITs) and stocks from Real Estate Holding
\& Development Companies. As there is no index to be found that is
geography specific for the real estate holdings in the study, we will
use a diversified European index for all real estate instruments.
\item As the liquidity of government bonds becomes thin with longer maturities,
we use the 10-years spreads for the 10- and 20-years bonds in the
study.
\end{itemize}
As the training of a neural network needs a lot of data, we use daily
data wherever possible. For some risk factors, single data points
are missing. We replace them with the preceding data point. If longer
data periods are missing for spreads or interest rates, we replace
this part with the time series of the same risk factor with a different
maturity. For the Ireland government bond spreads where there is a
four month time period where this is not possible, we use regression
and interpolation techniques with Portugal spreads which were at a
similar level at that time to fill in the gap.

These approximations in the risk factors will hold for the purpose
of this paper. If some of those risk factors are very important to
an insurance company which wants to adopt this concept, the data can
be either been sourced from a different data provider or some other
technique for data enrichment can be used, too.

High yield bond spreads are the only instruments where we have only
weekly and not daily data. There we use the rolling 12-month absolute
returns for all weekly data points and we interpolate between these
points using a regression from the BBB-spreads.

For not risk-free bonds, usually yields instead of spreads are available.
For this exercise, we transform them into spreads by subtracting the
relevant interest rate.

\section{\label{sec:Appendix-3-Optimization}Optimization of GAN architecture
using Wasserstein distance}

\selectlanguage{english}%
A full optimization is not possible as, see \citet[Chapter 2]{motwani2020novel},
the ``selection of the GAN model for a particular application is
a combinatorial exploding problem with a number of possible choices
and their orderings. It is computationally impossible for researchers
to explore the entire space.`` So, in our work, we\foreignlanguage{british}{
trained different GANs with}
\selectlanguage{british}%
\begin{itemize}
\item the number of layers for generator and discriminator varying between
2, 4, 6 and 8
\item the number of neurons for generator and discriminator varying between
100, 200 and 400
\end{itemize}
\selectlanguage{english}%
During the experiments, we fixed the following choices:
\selectlanguage{british}%
\begin{itemize}
\item Batch size is $M=200$
\item $k=10$ training iterations for the generator in each discriminator
training
\item Dimension of the latent space is 200, distribution of $Z$ is multivariate
normal with mean = 0 and std = 0.02
\item Initialization of generator and discriminator using multivariate normal
distribution with mean = 0 and std = 0.02
\item We use LeakyReLu with $\alpha=0.2$ as activation functions except
for the output layers which use Sigmoid (for discriminator) and linear
(for generator) activation functions. Additionally, we apply the regulation
technique batch normalization after each of the hidden layers in the
network.
\item We use the Adam optimizer with the parameters given in Section \ref{subsec:Implementation-of-a}
in Equation \ref{eq:adam}.
\end{itemize}
\selectlanguage{english}%
To arrive at one evaluation figure per GAN configuration, we aggregate
the 46 Wasserstein distances in each training iteration for each tested
GAN configuration $n$ by defining the following \emph{target function}

\[
tf_{n}=\min_{training\:iterations}\bigg(\max_{i=1,...,46}(W_{i}^{n})\bigg),\quad n=1,...,25
\]

with $W_{i}^{n}$ being the Wasserstein distance between \foreignlanguage{british}{the
empirical distribution functions of the training and the generated
data for risk factor} $i$ in GAN configuration $n$, as mentioned
in Section \foreignlanguage{british}{\ref{subsec:Implementation-of-a}}.
As in our case, we test 16 different configurations for the number
of layers and 9 different configurations for the number of neurons,
$n$ varies between 1 and 25.

First, we run the GAN with 16 different configurations of layers for
the discriminator $D$ and the generator $G$ with 200 neurons in
each layer of the network. For each neural network we use between
2 and 8 layers in steps of 2 layers. The resulting $tf_{n}-$values
are the following (the number of layer in $D$ are in the columns,
those of $G$ in the rows):
\begin{table}[H]
\begin{centering}
\begin{tabular}{ccccc}
\toprule 
number of layers in D/G &
2 &
4 &
6 &
8\tabularnewline
\midrule
\midrule 
2 &
0.143 &
0.252 &
0.636 &
0.799\tabularnewline
\midrule 
4 &
1.036 &
\textbf{0.118} &
0.235 &
0.435\tabularnewline
\midrule 
6 &
0.947 &
0.172 &
0.197 &
0.281\tabularnewline
\midrule 
8 &
0.807 &
0.188 &
0.171 &
\selectlanguage{british}%
0.178\selectlanguage{english}%
\tabularnewline
\bottomrule
\end{tabular}
\par\end{centering}
\caption{$tf_{n}$-values for varying number of layers in both networks}
\end{table}

The minimum of our target function $tf_{n}$ is reached for both neural
networks having 4 layers (marked in bold). We cannot confirm in this
experiment the thesis of \citet[p. 33]{goodfellow2016nips} stating
that the discriminator is usually deeper than the generator. What
we can clearly see is that the number of layers plays an important
role in the performance of the GAN.

We run the GAN with 9 different configurations of neurons per hidden
layer for the discriminator $D$ and the generator $G$. In this experiment,
we always set the number of neurons per layer being equal inside the
respective neural network. The resulting $tf_{n}-$values are the
following (the number of neurons per layer in $D$ are in the columns,
those of $G$ in the rows):
\begin{table}[H]
\begin{centering}
\begin{tabular}{cccc}
\toprule 
number of neurons per layer in D/G &
100 &
200 &
400\tabularnewline
\midrule
\midrule 
100 &
0.180 &
0.117 &
0.125\tabularnewline
\midrule 
200 &
0.185 &
0.118 &
\textbf{0.108}\tabularnewline
\midrule 
400 &
0.197 &
0.116 &
0.124\tabularnewline
\bottomrule
\end{tabular}
\par\end{centering}
\caption{$tf_{n}$-values for varying number of neurons per layer in both networks}
\end{table}

The minimum of our target function $tf_{n}$ is reached for the discriminator
having 400 and the generator having 200 neurons per layer (marked
in bold). This is the configuration which we then used in our further
research.\selectlanguage{british}%

\end{document}